\documentclass[10pt]{iopart}

\usepackage{bm}
\usepackage{amssymb}
\usepackage{eqnarray}
\usepackage{courier}
\usepackage{color}
\usepackage{url}
\usepackage{diagbox}
\usepackage{hyperref}

\usepackage{caption}
\usepackage[dvipsnames]{xcolor}

\usepackage{graphicx}
\usepackage{epstopdf}
\usepackage{threeparttable}

\expandafter\let\csname equation*\endcsname\relax
\expandafter\let\csname endequation*\endcsname\relax
\usepackage{amsmath}

\DeclareMathOperator*{\argmin}{arg\,min}

\definecolor{ao(english)}{rgb}{0.0, 0.5, 0.0}

\begin{document}

\title[Deep CAE for action potential compression]{Deep Compressive Autoencoder for Action Potential Compression in Large-Scale Neural Recording}

\author{Tong Wu$^1$, Wenfeng Zhao$^1$, Edward Keefer$^2$, and Zhi Yang$^{1*}$}

\address{$^1$ Biomedical Engineering, University of Minnesota, Minneapolis, MN, 55455, USA}
\address{$^2$ Nerves Incorporated, Dallas, TX, 75214, USA}

\ead{yang5029@umn.edu}
\vspace{10pt}

\begin{abstract}
\textit{Objective.}
Understanding the coordinated activity underlying brain computations requires large-scale, simultaneous recordings from distributed neuronal structures at a cellular-level resolution.
One major hurdle to design high-bandwidth, high-precision, large-scale neural interfaces lies in the formidable data streams (tens to hundreds of Gbps) that are generated by the recorder chip and need to be online transferred to a remote computer.
The data rates can require hundreds to thousands of I/O pads on the recorder chip and power consumption on the order of Watts for data streaming alone.
One of the solutions is to reduce the bandwidth of neural signals before transmission.
\textit{Approach.}
We developed a deep learning-based compression model to reduce the data rate of multichannel action potentials.
The proposed compression model is built upon a deep compressive autoencoder (CAE) with discrete latent embeddings.
The encoder network of CAE is equipped with residual transformations to extract representative features from spikes, which are mapped into the latent embedding space and updated via vector quantization (VQ).
The indexes of VQ codebook are further entropy coded as the compressed signals.
The decoder network reconstructs spike waveforms with high quality from the quantized latent embeddings through stacked deconvolution.
\textit{Main results.}
Extensive experimental results on both synthetic and in-vivo datasets show that the proposed model consistently outperforms conventional methods that utilize hand-crafted features and/or signal-agnostic transformations and compressive sensing by achieving much higher compression ratios (20--500$\times$) and better or comparable reconstruction accuracies.
{Testing results also indicate that CAE is robust against a diverse range of imperfections, such as waveform variation and spike misalignment, and has minor influence on spike sorting accuracy.}
Furthermore, we have estimated the hardware cost and real-time performance of CAE and shown that it could support thousands of recording channels simultaneously without excessive power/heat dissipation.
\textit{Significance.}
The proposed model can reduce the required data transmission bandwidth in large-scale recording experiments and maintain good signal qualities, which will be helpful to design power-efficient and lightweight wireless neural interfaces.
The code of this work has been made available at \url{https://github.com/tong-wu-umn/spike-compression-autoencoder}.
\end{abstract}

\clearpage

%
%
%
\ioptwocol

\section{Introduction}
Understanding the coordinated activity underlying brain computations requires large-scale, simultaneous electrophysiological recordings from distributed neuronal structures at a cellular-level resolution.
There is a recent trend to develop high-density neural interfaces that include tens of thousands and even hundreds of thousands channels \cite{insel2013nih,darpa2016nesd}. 
For example, multiple studies have been proposed that developed high-channel-count, high-precision neural recorders \cite{berenyi2014large,yang2016embc} and high-density microelectrode arrays \cite{jun2017fully,tsai2015high}.
Given the successful development of high-density arrays, it requires streaming the data to a remote computer for processing, which can be challenging:
large-scale recording experiments can produce data at tens of hundreds of Gbps \cite{stevenson2011advances}, which would require hundreds to thousands of I/O pads on the recorder chip and power consumption on the order of Watts for data streaming alone.
To solve the problem, it requires to compress the neural data in the recorder chip before transmission, and reconstruct the data or directly utilize the compressed signals on the remote computer.

Spike detection can be used as a method to compress high frequency neural data, in which the spike waveforms or only timestamps are transmitted \cite{Karkare-JSSC-2011,7055372,wu2015nonparametric,Gosselin-TBCAS-2009}.
The compression ratio (CR) achieved by this type of methods depends on both neuronal firing rate and signal-to-noise ratio (SNR): 
when the SNR is poor, it requires transmitting all likely events that increases the firing rates;
firing rate can also be high during some neural dynamic states such as bursting;
in both cases, CR would drop rapidly.
Feature extraction methods can be used to represent spikes with a few hand-engineered features (e.g., durations of depolarization and hyperpolarization), and achieve high CR.
A limitation is that raw spike waveforms, which experimentalists may want to keep, cannot be restored from the extracted features.
Besides, there have been few works on efficient on-chip implementation of these methods.
To further boost CR without losing raw waveforms, spike compression is widely adopted and often approached in the context of lossy data compression, which transforms spikes into a different domain where a reduced number of coefficients can be identified and quantized to represent spikes more efficiently.
In this scenario one must balance two competing costs: the count of discretized representation (rate) and the error arising from quantization (distortion) \cite{balle2016end}, i.e., the rate-distortion trade-off.
Frequently used transformations include principal component analysis (PCA) \cite{wu2017streaming,chae2008128}, discrete cosine transform (DCT) \cite{shinde2011comparison}, linear approximations \cite{zordan2014performance},  and wavelet transform \cite{shaeri2015method,hosseini2014data}, etc.
These transformations are either nonparametric, or in which parameters can be learned from data.
However, {due to their signal-agnostic natures and that they are not explicitly optimized for compression}, features exposed in the transformed domains often lack representational power, resulting in limited CR.
A popular strategy emerged in recent years that can be used to compress electrophysiological data is compressive sensing \cite{zhao2018chip,zhang2015closed,zhang2016communication}.
Compressive sensing can provide CR comparable to transformation-based approaches with much simpler computational complexities for data encoding, leaving most of the computational burdens to the remote processor \cite{sun2017training}.

Another type of approaches that can potentially boost CR is learning-based compression, such as vector quantization (VQ), where a signal-dependent codebook is learned from data and only the indexes of individual codeword in the codebook are transmitted \cite{craciun2011wireless}.
{This is different from conventional compression techniques that exploit efficient coding of the bit patterns of data after quantization;
rather, it seeks ``distilled'' representations of the information content of signals, which could be more advantageous to achieving higher CR for data subject to certain statistical distributions.}
However, distributions of real-world data are usually in high-dimensional space, and are difficult or even intractable to estimate analytically.
Furthermore, learned codebooks often ``overfit'' training data and do not generalize well, which leads to the requirement of frequent re-training and transmitting the entire codebook or uncompressed data that may interrupt data transmission and deteriorate CR.
To make the learning-based compression approach effective and practical in large-scale neural recording, we need to address the following issues:
(i) The size of the codebook cannot grow arbitrarily large to maintain good signal reconstruction accuracy in situations of low SNRs and/or diverse spike waveforms;
(ii) The codebook must represent inherent spike features, such that the compression algorithm can be robust to non-stationarities of neural activities, e.g., waveform variations.
These requirements entail the search of the ``optimal codebook'' that best characterizes the statistics of spikes by taking a reasonable amount of samples from spike data manifold in a low-dimensional feature space.

We propose to construct high-quality codebooks using deep neural network (DNN) to facilitate effective learning-based compression.
DNN-based feature extraction relies heavily on carefully designed network architectures and well tuned hyperparameters that are tailored for specific types of data.
Hence it is crucial to design DNN structures that are suitable to extract representative features from multichannel spikes.
Another challenge is the integration of a DNN model, normally with millions of parameters, into a neural recorder chip with limited hardware resources and power budgets.
This requires hardware-aware design optimizations to obtain an extremely efficient and compact DNN model that is feasible for on-chip implementation without compromising performance.

In this paper, we tackle these challenges by proposing a lightweight DNN model -- compressive autoencoder (CAE) -- that can compress thousands of spikes simultaneously by 20--500$\times$ with signal quality comparable to or better than that of existing approaches.
Our main contributions include:
{(i) Instead of hand-crafted features or signal-agnostic transformations, we use convolutional neural network (CNN) along with vector quantization to extract and sample hierarchical features, which exhibit strong representational capability and generalize well to unseen spikes;}
(ii) We show that CAE is capable of leveraging geometrical information of spikes from multichannel recording, which is useful to expose localized features and improve qualities of reconstructed signals;
{(iii) We demonstrate that CAE can allow for high compression ratios without noticeably compromising spike sorting accuracy;}
(iv) CAE features an asymmetric model structure for signal encoding and decoding, where the encoding part (along with quantization) requires fewer than 20K parameters, which is over 40$\times$ smaller than the decoding and suitable for efficient on-chip implementation into large-scale neural recording systems.

The rest of the paper is organized as follows.
Section 2 describes the system architecture of the proposed spike compression model and its theoretical basis.
{Section 3 and 4 presents and discusses the experimental results.
Section 5 concludes the paper.}

\section{Methods}

\subsection{Compressive autoencoder for neural data compression}
At the core of the proposed model is autoencoder \cite{bengio2009learning}, a neural network structure widely used to learn compact data representations by forcing outputs to be identical as inputs and imposing constraints in the latent space.
Mathematically, the general operation of an autoencoder can be described as $\hat{\mathbf{x}}=g_s(g_a(\mathbf{x}; \phi); \theta)$, where {$\mathbf{x}$ and $\mathbf{\hat{x}}$ are input and output data;} $g_a$ and $g_s$ denote \emph{analysis} and \emph{synthesis}, respectively, or are commonly referred to as encoder and decoder (parameterized by $\phi$ and $\theta$).

In the context of lossy data compression, the operation of CAE becomes:
\begin{align}
\label{define}
\phi, \theta & =\argmin_{\phi, \theta}\|\mathbf{x}-\mathbf{\hat{x}}\|_2^{2}, \nonumber \\
& \text{subject to} \left\{\begin{array}{rcl}
\mathbf{y} &=& g_a(\mathbf{x}; \phi) \\
\mathbf{\hat{y}} &=& \text{quantize}(\mathbf{y}) \\
\mathbf{\hat{x}} &=& g_s(\mathbf{\hat{y}}; \theta)
\end{array}\right.
\end{align}
where the \texttt{quantize} function discretizes encoder outputs and introduces quantization error.
{Conventionally, the loss function of a CAE that optimizes both bit rates and distortion is}:
\begin{equation}
\label{rdloss}
\mathcal{L}_{CAE} = \underbrace{-\text{log}_2Q(\mathbf{\hat{y}})}_{\text{Number of bits}} \quad + \quad \alpha \cdot \underbrace{d(\mathbf{x}, \mathbf{\hat{x}})}_{\text{Distortion}},
\end{equation}
where $Q(\cdot)$ is the operation to estimate the discrete probability distribution of discretized data, and $\alpha$ is used to adjust the rate-distortion trade-off.

\begin{figure*}[t]
\center
\includegraphics[width=\linewidth]{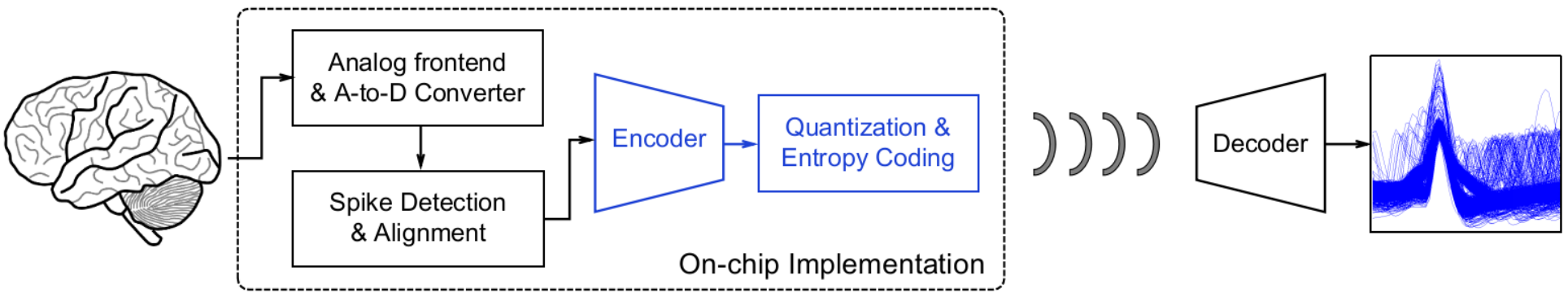}
\caption{Conceptual diagram of deploying CAE in wireless neural recording system. For simplicity, microelectrodes are omitted from the figure. Spike snapshots are from synthetic datasets \texttt{\texttt{Wave\_Clus}} \cite{Quiroga-NeuralComput-2004}.}
\label{diagram}
\vspace{-10pt}
\end{figure*}

The conceptual diagram of deploying CAE into wireless neural recording systems is illustrated in Figure \ref{diagram}.
After recorded from analog frontend circuitry and digitized, spikes are extracted from raw recording data and aligned.
For CAE, only the encoder and the quantization block need to be on-chip implemented; the decoder can run on a remote computer.
$\theta$ and $\phi$ are programmable to allow flexible choices of rate-distortion trade-offs.
The indexes of VQ codebook corresponding to encoder outputs are coded and transmitted, leading to significant data rate reduction.

In practice, direct optimization of CAE using equation (\ref{rdloss}) proves difficult, because (i) $Q(\cdot)$ and \texttt{quantize} are typically non-differentiable thus cannot be updated via back-propagation, and (ii) joint optimization of both rate and distortion requires complex computations and carefully designed training schedules.
For example, in \cite{theis2017lossy} an extra Gaussian scale mixture model is used to model distribution of coefficients and estimate bit rates, as well as the requirement of fine-tuning a pre-trained autoencoder for different bit rates.
On top of that, \cite{theis2017lossy} needs an incremental training that gradually ``releases'' coefficients for updates and takes up to 10$^6$ iterations to achieve good performance.
The first difficulty can be solved by directly copying the gradient of decoder inputs to the encoder outputs during training, bypassing the quantization block \cite{balle2016end,theis2017lossy,NIPS2017_7210}.
To address the second difficulty, we remove the rate penalty from equation (\ref{rdloss}) and leave the size of VQ codebook programmable by the users.
{The advantage of this modification is two-fold:
(i) We found that penalizing only distortions can lead to fast convergence of training (typically $\sim$200 epochs) with good performance.
This is beneficial to fast and simplified deployment of the model onto mobile hardware platform for real-time spike compression;
(ii) It allows straightforward optimization towards lower distortion.
To update VQ codebook, we add the Euclidean distance between encoder outputs and corresponding VQ codewords into the overall loss function.
{After the modification, the loss function of the proposed CAE now becomes:}
\begin{equation}
\mathcal{L}_{CAE} = d(\mathbf{x}, \mathbf{\hat{x}}) + d(\mathbf{y}, \mathbf{\hat{y}}).
\end{equation}
{Compared with equation (2), the rate penalty is removed and another Euclidean loss to optimize VQ codewords is added.}
Thus the VQ codewords can be updated in the same way as other parameters via back-propagation, which simplifies the network training.
}

\subsection{Encoder and decoder networks}
In this work, we designed the encoder and decoder networks based on the structures of deep CNN to extract features from spikes for compression/reconstruction.

{
Fusing spatial and temporal information by stacking a set of convolutional filters interleaved with non-linearity and pooling is essential to enhance the representational power of DNN \cite{hu2017squeeze}.
The extraction and fusion of spatial features is realized within the computation of each CNN layer.
For a CNN layer with $C_{in}$ input channels and $C_{out}$ output channels, the value of the $j_{th}$ output channel can be described as
\begin{align}
\text{out}(N_i, C_{out_j}) = \sum_{k=0}^{C_{in}-1}\text{weight}(C_{out_j}, k) \star \text{input}(N_i, k),
\label{cnn}
\end{align}
where $N$ is the batch size, \texttt{weight} is the coefficients of a CNN filter, $\star$ denotes cross-convolution, i.e., $I \star K(i, j)=\sum_m\sum_nI(i+m, j+n)K(m, n)$.
Thus it is clear that to compute one channel output of a CNN layer requires all $C_{in}$ input channels, which essentially realizes fusion of spatial features from previous layers and propagation of the features to subsequent layers.
}

{
The temporal features of spikes are implicitly extracted by enforcing a local invariance of CNN filters, i.e., the coefficients of each filter stay the same when convolving with the entire input sequence along the temporal direction.
The underlying assumption is that there exists hidden temporal structures that remain invariant and are common to diverse spike morphologies.
Eventually the temporal features are approximated by VQ codewords.
}

In general, deep networks are representationally superior to shallow ones to allow for sufficient information fusion and propagation \cite{cohen2018boosting}.
However, there are two major constraints with deeper networks in our application.
First, with more layers, the amount of network parameters (weights of convolutional filters, etc.) increases drastically, making it more difficult for on-chip implementation in the neural recorder chip.
Second, as in large-scale recording the probes can span a relatively large brain area or a long range of vertical structures, the underlying signal characteristics of neural data from adjacent recording sites in one local region will be different from that of other regions due to the differences in current path between the neurons and the electrodes.
Therefore, features learned by the convolutional filters from all channels may not be optimally representative for each distributed recording regions, especially when one cannot afford sufficiently deep networks on-chip.

\begin{figure*}[t]
\centering
\includegraphics[width=.95\linewidth]{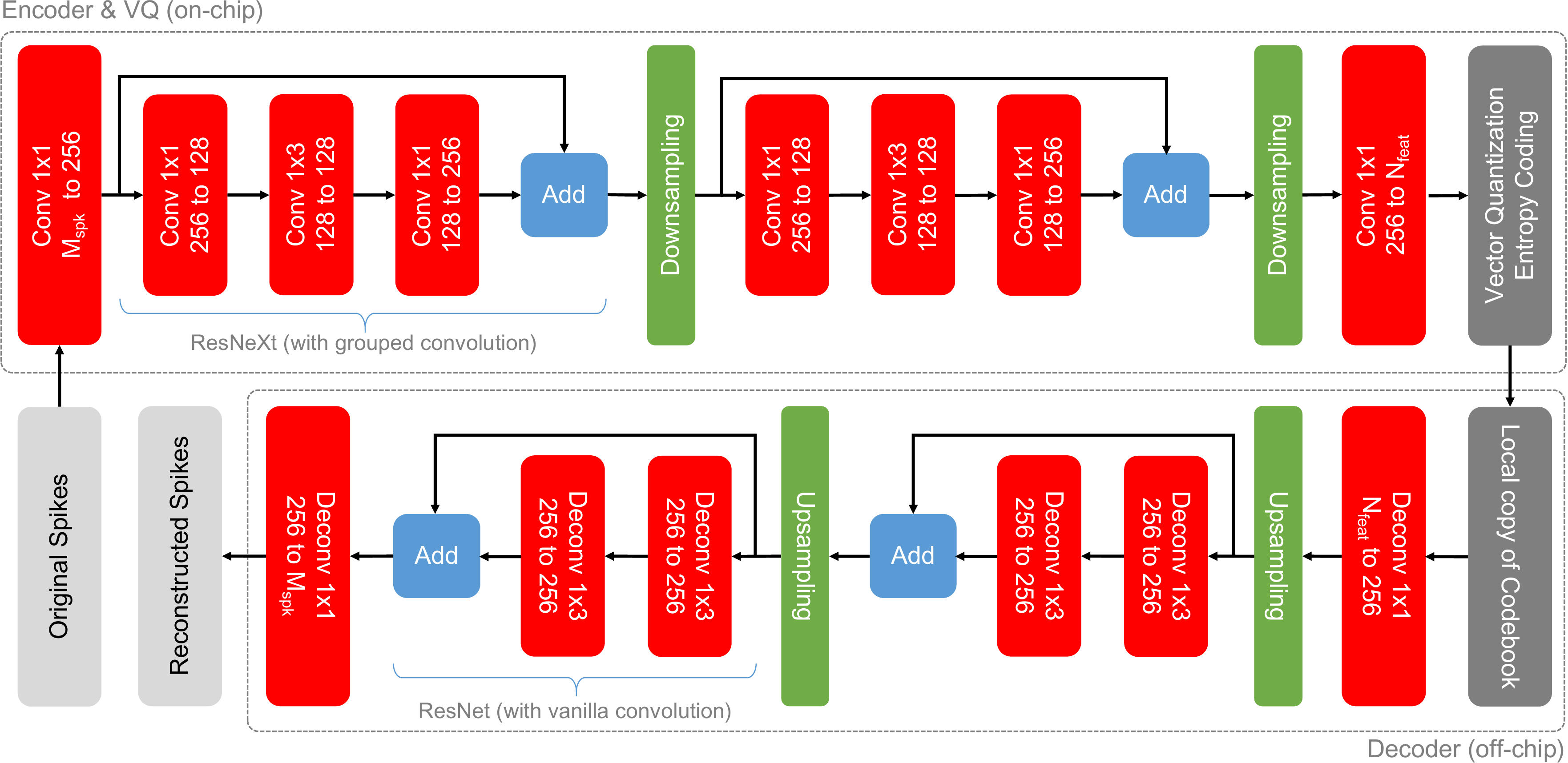}
\caption{{Structural diagram of the proposed spike compression model. Activation layers and normalization layers are skipped from the diagram for simplicity.}}
\label{cae}
\vspace{-5pt}
\end{figure*}

To circumvent the first limitation, the main structures of the encoder and decoder are based on \texttt{\texttt{ResNet}}, i.e., residual network with identity-mapping shortcut connections \cite{he2016deep}.
It has the capability of representing a richer set of complex features compared with other network structures with the same or even larger model size, presumably thanks to its behavior like ensembles of relatively shallow networks by introducing the shortcut connection \cite{veit2016residual}.
To resolve the second constraint, more effective ways of organizing convolutional filters are explored.
One promising configuration is grouped convolution \cite{krizhevsky2012imagenet}.
In vanilla convolution, {a total number of $C_{in} \times C_{out}$ filters are required for a convolutional layer as evidenced in equation (\ref{cnn}).}
With grouped convolution, the input and output channels are evenly split into $C$ groups, thus the number of filters is reduced $C$-fold.
By restricting the ``receptive fields'' of convolutional filters, local features can be preserved within each group and are more representative of neural spikes recorded from corresponding physical channels.
Moreover, the amount of parameters can be significantly reduced thus more hardware efficient.
In this work, 32 groups are used in all \texttt{ResNeXt} modules (following the naming in \cite{xie2017aggregated}, i.e., \texttt{ResNet} with grouped convolution).

The detailed diagram of the proposed spike compression network is shown in Figure \ref{cae}.
In the encoder network, the input convolutional layer with kernel size 1$\times$1 maps detected spikes organized in $M_{spk}$ channels to a 256-channel feature space.
Following the first channel-expansion layer, we cascade two stages of \texttt{ResNeXt} to enhance the feature extraction capability.
The main pathway of each \texttt{ResNeXt} is configured in bottleneck connection \cite{he2016deep}, consisting of a stack of 3 layers with 1$\times$1, 1$\times$3, and 1$\times$1 convolutional kernels, respectively, where the 1$\times$1 layers are responsible for reducing and restoring dimensions, and the 1$\times$3 layer extracts features with halved input/output dimensions.
Each \texttt{ResNeXt} is followed by a 2$\times$ downsampling along the temporal dimension.
The last stage of the encoder network is a vanilla 1$\times$1 convolutional layer that aggregates the features learned from previous stages and reduces the channel dimension from 256 to $N_{feat}$.
The decoder network is a reverse implementation of its encoder counterpart, where convolution and downsampling are replaced by transposed convolution (deconvolution) and upsampling.
As the decoder network is implemented on a remote computer with abundant computational resources and is primarily used to reconstruct the inputs, we stack two 1$\times$3 deconvolutional layers in the main pathway of each \texttt{ResNet} instead of using grouped convolution to enhance the reconstruction capability.

\subsection{Dimensionality reduction, vector quantization, and entropy coding}
For $M_{spk}$ input spikes in $D$-dimension, the encoder outputs $N_{feat}$ feature vectors in $d$-dimension.
The operation of \texttt{quantize} can be described as $R^d \rightarrow C$ that maps a feature vector in $d$-dimensional space into a codebook $C$ containing $K$ codewords $\{C_i; i=1, 2, ..., K\}$.
Each codeword requires $\text{log}_2K$ bits to represent in unsigned binary representation.
Therefore, the overall data rate reduction without entropy coding is 
\begin{equation}
\text{CR}=\frac{M_{spk} \cdot D \cdot W}{N_{feat} \cdot \text{log}_2K},
\label{cr}
\end{equation}
where $W$ is the original bit-length of spike sample, typically 10--16 bits; 
\emph{D} is the original spike dimension, typically 40--80.
$d$ is not involved in the denominator of CR as only the codeword indexes need to be transmitted.
Note that in the rest of paper, the reported CRs are calculated using the entropy of codewords.
The actual CR will be slightly decreased due to the overhead of using a practical code (e.g., Huffman coding).

In this work, we adopt a Voronoi vector quantizer that partitions the CAE latent space into $K$ cells, of which the centroids are the codewords.
Each cell consists of all points $\mathbf{y}$ which have less distortion when reproduced with codeword $\mathbf{\hat{y}}$ than with any other codeword.
All codewords are initialized following a uniform distribution and updated according to the distance between the current values of codewords and the feature vectors output by the encoder.

{
Finally we discuss the distortion introduced by quantization in the CAE latent space.
Given a $d$-dimensional quantizer with a distortion measure $\|x-y\|^r$ ($r\ge1$), we have a high rate lower bound of the quantization distortion as (\cite{gray2012source}):
\begin{align}
Dist. &\ge \frac{d}{d+r}(V_d)^{-r/d}e^{-\frac{r}{d}(H(Q(X))-h(X))} \nonumber \\
& \approx \frac{d}{d+r}(V_d)^{-r/d}e^{-\frac{r}{d}(\text{log}_2K-h(X))} \nonumber \\
& = \frac{d}{d+r}(V_d \cdot K)^{-r/d}e^{\frac{r}{d} \cdot h(X)},
\end{align}
where $Q(X)$ is the entropy of quantizer output, $h(X)$ is the differential entropy of quantizer input $X$, $V_d$ is the volume of unit sphere in $d$-dimensional space.
In practice, the probability density function of VQ codewords is approximately uniform, thus $H(Q(X))$ is very close to $\text{log}_2K$.
Hence for a CAE model with $d$, $r$, and $K$ fixed, the best performance (lowest distortion) depends primarily on the complexity of input, which is approximated as $h(X)$.
}

\subsection{Parameter configuration and model training}
As suggested in equation (\ref{cr}), parameters $M_{spk}$, $N_{feat}$, and $K$ jointly determine the achievable CR ($D$ and $W$ are determined by the recording specification thus excluded from discussion).

The choices of these parameters require careful trade-offs between CR, signal quality, and hardware cost.
In general, a larger codebook is needed for the compression of noisier spikes or spikes recorded from many channels and with more diverse waveforms to ensure the reconstruction accuracy, as evidenced in equation (6).
The ratio $M_{spk}/N_{feat}$ affects the trade-off between CR and reconstruction accuracy, and their actual values have little impact on the performance.
However, larger $M_{spk}$ and $N_{feat}$ would lead to increased hardware cost.
$M_{spk}$ is empirically set as 4 in all the experiments (except for Section 3.3.4).
For recordings from more than 4 channels, spikes from adjacent channels are grouped together and sent into one of the $M_{spk}$ ports.

The design and testing environment is Intel i7-6800K@3.40GHz, NVIDIA GeForce Titan Xp 12GB, 32GB memory, 256GB SSD, and Ubuntu 16.04 LTS.
The proposed CAE model is implemented using the deep learning framework PyTorch 0.4.1 (with CUDA 9.0) \cite{paszke2017automatic}.
We used the ADAM optimizer \cite{kingma2014adam} with learning rate 1e-3 and evaluated the model performance after 500 epochs with batch-size 48 in all the experiments.

\subsection{Datasets preparation}
The synthetic dataset we have chosen is \texttt{Wave\_Clus} from University of Leicester \cite{Quiroga-NeuralComput-2004}, which has been widely used in the evaluation of spike sorting algorithms.
The dataset is generated by adding several spike waveform templates to background noise of various levels, thus realizing different SNRs.
We used four datasets \texttt{C\_Easy1\_noise01}, \texttt{C\_Easy2\_noise01}, \texttt{C\_Difficult1\_noise01}, \texttt{C\_Difficult2\_noise01}, each of which was generated using different spike templates.
We used the ground truth spike times included in the datasets to extract spikes from the continuous data.
Spikes from the four datasets were grouped together, presenting more challenge for compression due to combined spike templates.
All spikes were aligned to their maximum peaks with 64 samples per spike.

The first in-vivo dataset we used is the data recorded from the rat CA1 hippocampal region using tetrodes that are publicly available as \texttt{HC1} \cite{harris2000accuracy,Henze-JNeuronphy-2000}.
The tetrodes consist of four 13$\mu$m nichrome wires bound together by twisting them and melting their insulation \cite{gray1995tetrodes}.
The dataset \texttt{d53301} was used for evaluation, which contains four extracellular channels from tetrodes and one intracellular channel from a micropipette.
Extracellular signals were high-pass filtered at 300Hz.
Spikes were extracted from the four extracellular channels with the spike times determined by the occurrences of intracellular action potentials on the micropipette.
All extracted spikes were aligned to their maximum peaks with 48 samples per spike.

To test the performance of the proposed method in compressing neural signals from more recent large-scale, high-density recording setups, we used the in-vivo data recorded from an awake, head-fixed mouse using the \texttt{Neuropixels} probe \cite{lopez201622}.
\texttt{Neuropixels} has 384 recording sites with 70$\times$20$\mu$m$^2$ per site.
The neural data are band-pass filtered at 300--5000Hz.
Spikes were detected from each channel by amplitude thresholding.
The threshold was set at
\begin{equation}
\text{Threshold}=5 \times \text{median}\{\frac{|X|}{0.6745}\}
\end{equation}
where $X$ is the band-pass filtered signal \cite{Quiroga-NeuralComput-2004}.
All detected spikes were aligned to their maximum peaks with 48 samples per spike.
All channels used the same parameter setting for detection and no further fine-tuning was performed.
Therefore the detected spikes contain a large number of false alarms contributed by various noise sources.
The existence of many non-spike activities would significantly increase the difficulty of spike compression due to the diverse signal and noise characteristics.
A successful compressor should reduce the bandwidth of both spikes and non-spike activities at the same time and shift the burden of carefully differentiating spikes from noise to a remote computer.
 
We used the mean squared error (MSE) to optimize the neural network.
To measure the signal reconstruction accuracy and also allow for comparison with other works, we reported accuracy in average signal to noise and distortion ratio (SNDR) defined as:
\begin{equation}
\text{SNDR}=20 \cdot \text{log}_{10}\frac{\|X\|_2}{\|X-\hat{X}\|_2}.
\end{equation}

The bit-length of all spike data is assumed 16-bit, which is a common setting adopted in commercial neural recording devices.

\subsection{Methods for comparison}
We have chosen three transformation-based methods for comparison, including PCA, DCT, and discrete wavelet transform (DWT).
We also compared with a recent work based on compressive sensing, group weighted analysis $l_1$-minimization (GWALM) \cite{sun2017training}, that showed better performance compared with other compressive sensing-based approaches.

\subsubsection{Proposed CAE}
For each dataset, spike data are randomly split into two parts for training (50\%) and testing (50\%).
The random training/testing partition was repeated five times on each dataset and for each method we took the average performance as the final results.
We train the network using the training data and evaluate the model performance on the testing data.
The number of partitions $K$ in VQ is assumed powers of two.
The CR is computed using equation (\ref{cr}) with the $\text{log}_2K$ replaced by the entropy of the codeword indexes on the testing data.

\subsubsection{PCA}
We apply PCA on the training spikes, and keep the leading $m$ eigenvectors as the transformation matrix.
For compression, we multiply the transformation matrix with the testing spikes to obtain principal components as the compressed signals.
The number of eigenvectors $m$ is set as 2, 4, 6, 8, 10 for \texttt{Neuropixels}; 1, 2, 4, 6, 8 for \texttt{HC1} and \texttt{Wave\_Clus}.
Each principal component is represented using the same bit-length as that of spikes.
The CR can be computed as $\text{CR}=D / m$.

\subsubsection{DWT}
Spikes are first transformed into wavelet representation and the $m$ most significant coefficients are kept (others are zeroed).
The number of wavelet coefficient $m$ is set from 2 to 12 with an increment of 2 for all datasets.
The $Symmlet4$ wavelet basis is used as it is advantageous over other wavelet basis families for processing neural signals \cite{oweiss2007scalable}.
Each wavelet coefficient is represented using the same bit-length as that of spikes.
The CR can be computed as $\text{CR} = D \cdot W / (W \cdot m + D)$, where the $D$ bits in the denominator are used to denote the positions of the $m$ non-zero coefficients.

\subsubsection{DCT}
We keep the $m$ leading coefficients of each spike after transformed by DCT.
$m$ is set from 8 to 16 with an increment of 2 for \texttt{Wave\_Clus}; 6, 8, 10, 11, 12 for \texttt{Neuropixels}; from 6 to 10 with an increment of 1 for \texttt{HC1}.
Each coefficient is represented using the same bit-length as that of spikes.
The CR can be computed as $\text{CR}=D / m$.

\subsubsection{GWALM}
First, an analysis model is adopted to enforce sparsity of spikes;
second, a multi-fractional-order difference matrix is constructed as the analysis operator;
third, by exploiting the statistical properties of the analysis coefficients, a group weighting approach is developed to enhance the performance of analysis $l_1$-minimization.
Each spike was compressed to a vector of length $m$. 
$m$ is set from 8 to 16 with an increment of 2 for \texttt{Neuropixels}; from 8 to 12 with an increment of 1 for \texttt{HC1}; from 12 to 17 with an increment of 1 for \texttt{Wave\_Clus}.
The CR can be computed as $\text{CR}=D/m$.
More details of the algorithm can be found in \cite{sun2017training}.

\begin{figure*}[t]
\center
\includegraphics[width=\linewidth]{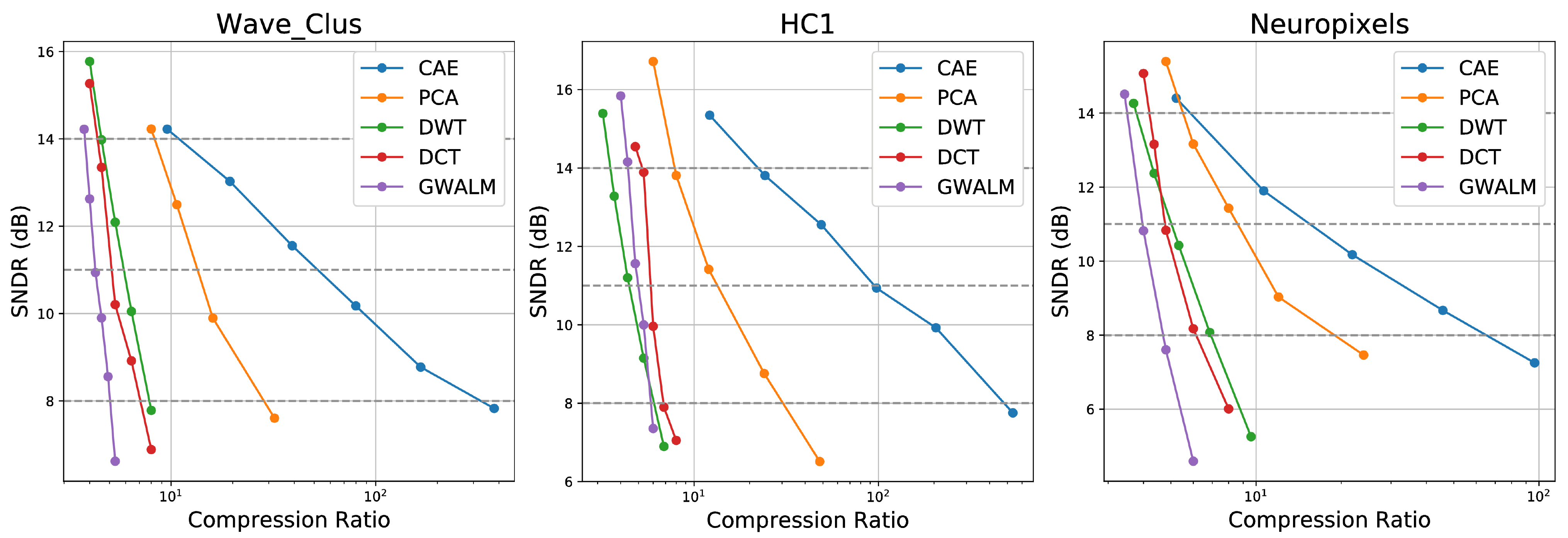}
\caption{{Rate-quality curves of all methods on \texttt{Wave\_Clus}, \texttt{HC1}, and \texttt{Neuropixels} datasets. Vertical axis is signal reconstruction accuracy measured in SNDR. Horizontal axis is compression ratio (also in logarithmic scale). Configuration of CAE: $M_{spk}/N_{feat}=1/4$, $K=128, 32, 512$ for \texttt{Wave\_Clus}, \texttt{HC1}, and \texttt{Neuropixels} datasets, respectively. SNDRs at 8dB, 11dB, and 14dB are highlighted in gray lines.}}
\label{rate_distort}
\vspace{-10pt}
\end{figure*}

\section{Results}
\subsection{Compression of synthetic and in-vivo spikes}
We run CAE and other approaches on each of the synthetic and in-vivo datasets.
The rate-quality curves of all methods on each dataset are plotted in Figure \ref{rate_distort}.
Both horizontal and vertical axes are in logarithmic scale to clearly distinguish curves corresponding to different methods.
As shown in the figure, CAE outperforms all other methods, primarily by extending CR into the range of 20--500$\times$.
We also highlight three levels of reconstructed signal qualities measured in SNDR: 8dB, 11dB, and 14dB.
It is clear that CAE achieves much higher CRs on both synthetic and in-vivo datasets than other methods especially at SNDR of 8dB and 11dB, e.g., up to 500$\times$ CR on \texttt{HC1}, which is 15$\times$ better than PCA and over 70$\times$ better than DWT, DCT, and GWALM.
The performance gap on \texttt{Neuropixels} at SNDR of 14dB is small due to the more complex signal characteristics of spikes from hundreds of recording channels.
The qualities of the reconstructed spikes using CAE are illustrated in Figure \ref{recon_spks} and Figure \ref{spk_high_cr}, each of which shows 24 reconstructed spikes with various shapes randomly chosen from each dataset.

DWT, DCT, and GWALM have similar performances: their signal qualities decrease radically to smaller than 5dB when CR approaches 10$\times$.
PCA achieves better results than other conventional methods, possibly due to that in all linear projections, PCA can achieve the minimum reconstruction error given fixed input/output dimensions, which is consistent with our previous research \cite{wu2017streaming}.
{
It is worth noting that PCA can be considered as a type of linearized autoencoder optimized over MSE-based loss functions, which is similar to CAE; 
however, PCA cannot take advantage of nonlinear features that are representative for many high-dimensional data, thus its representation capability is inherently inferior to CAE.
}
Another limiting factor of PCA for compression is that its CR cannot exceed the original spike dimension as at least one principal component is required to represent one spike.
The performance of GWALM is not as good as others.

\subsection{{Evaluation of generalization capability of CAE}}
One common issue of learning-based compression methods is the generalization capability of codebooks learned from training data.
The compression performance is largely dependent on the similarity of the statistics of testing data relative to that of training data.
Thus it is crucial to extract representative features from training data that can capture the underlying statistical distributions as accurately as possible.
Such capability is of critical importance for a compression method to stay robust against various non-stationarities of neural activities.
For example, individual spikes in a burst can have more than 50\% amplitude variation according to simultaneous intracellular and extracellular recordings \cite{Henze-JNeuronphy-2000}.
Electrode drift is another common source that causes systematic changes in the shape and amplitude of recorded spike waveforms \cite{shan2017model}.

\begin{figure*}[t]
\center
\includegraphics[width=0.75\linewidth]{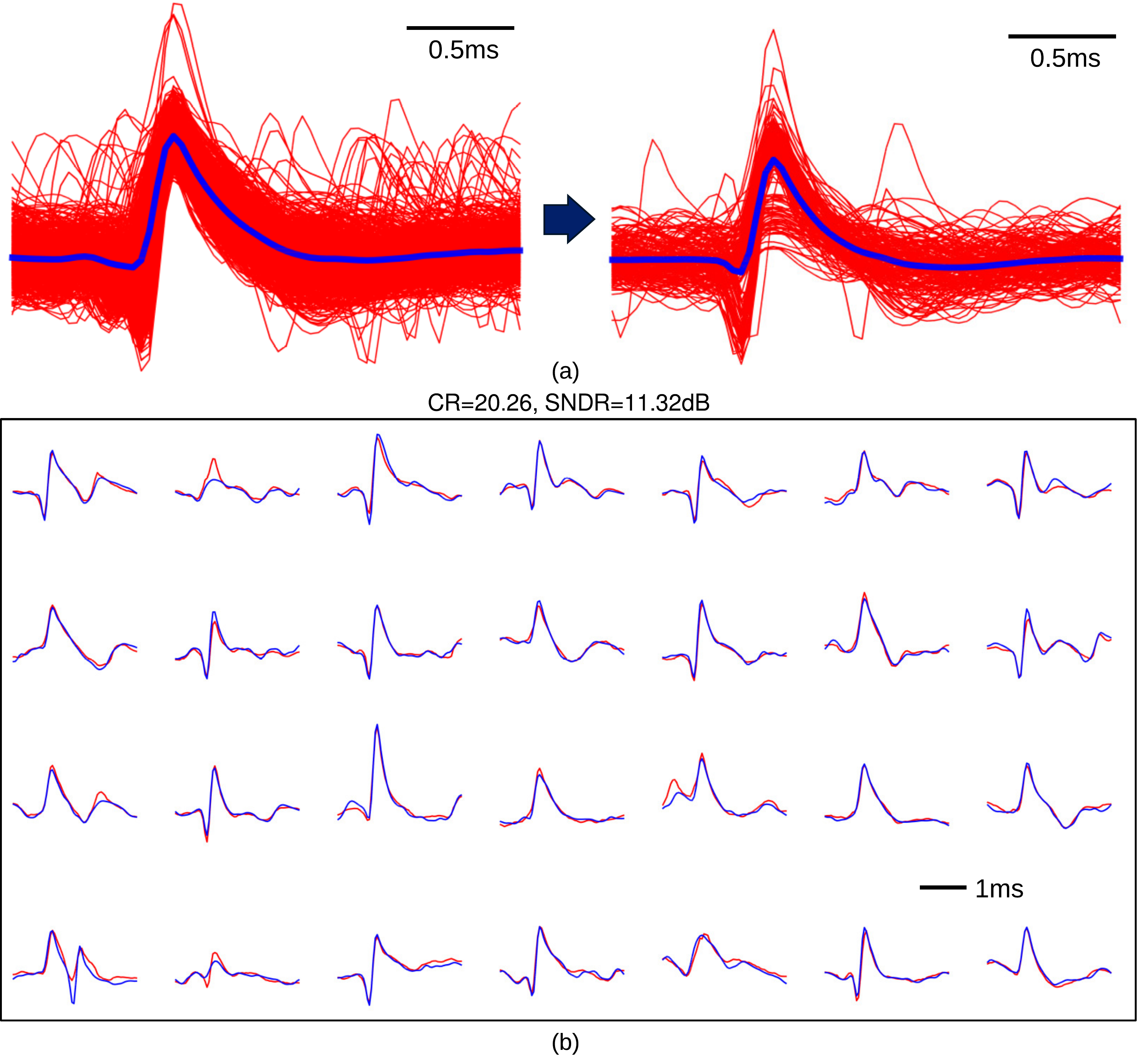}
\caption{(a) Left: the first 500 spikes in \texttt{C\_Drift\_Easy2\_noise015}; right: the last 200 spikes in the same sequence. Blue curves in each sub-figure represent the mean of all spikes. (b) Reconstruction of spikes in \texttt{C\_Drift\_Easy2\_noise015} using CAE. 28 spikes are randomly chosen from the last 200 spikes. Blue and red waveforms are original and reconstructed spikes, respectively. Configuration of CAE: $M_{spk}/N_{feat}=1/8$, $K=128$. Performance on testing dataset: CR=20.26, SNDR=11.32dB.}
\label{drift}
\vspace{-10pt}
\end{figure*}

\begin{table*}[t]
\centering
\caption{{Evaluation of CAE performance by using different sequences for training and testing.}}
\label{cae_train}
\begin{threeparttable}
\begin{tabular}{lcccc}
\hline
\multicolumn{1}{c}{\diagbox{Training}{Testing}} & \begin{tabular}[c]{@{}c@{}}C\_Difficult1\\ Noise 0.05\end{tabular} & \begin{tabular}[c]{@{}c@{}}C\_Difficult1\\ Noise 0.2\end{tabular} & \begin{tabular}[c]{@{}c@{}}C\_Difficult2\\ Noise 0.05\end{tabular} & \begin{tabular}[c]{@{}c@{}}C\_Difficult2\\ Noise 0.2\end{tabular} \\ \hline
C\_Difficult1 Noise 0.05 & \textcolor{Gray}{13.8223} & 8.271 & 12.2762 & 8.0851 \\
C\_Difficult1 Noise 0.2  & 13.0852 & \textcolor{Gray}{10.2059} & 14.4029 & 10.4476 \\
C\_Difficult2 Noise 0.05 & 11.6319 & 8.4129 & \textcolor{Gray}{15.8518} & 9.1695 \\
C\_Difficult2 Noise 0.2  & 12.3602 & 9.9329 & 15.2567 & \textcolor{Gray}{10.9564} \\ \hline
\end{tabular}
\begin{tablenotes}
\footnotesize
\item Configuration of CAE: $K=128$, $M_{spk}/N_{feat}=1/4$. Numbers are SNDR (dB).
\end{tablenotes}
\end{threeparttable}
\vspace{-10pt}
\end{table*}

To demonstrate that CAE indeed learns representative features instead of simply ``memorizing'' instances of training data, we used a synthetic dataset \texttt{C\_Drift\_Easy2\_noise015} from \texttt{Wave\_Clus} that simulates the effect of electrode drift and caused waveform variation.
The sequence contains 3444 spikes, and the shapes of spikes gradually change along the temporal axis.
We trained a CAE model using the first 500 spikes and tested the model on the last 200 spikes in the sequence.
Figure \ref{drift}(a) shows clear differences between the training and testing spikes, primarily including
(i) decrease of average spike amplitude,
(ii) reduced noise in the non-polarization parts of spikes, and
(iii) a newly emerged spike cluster with much smaller amplitude,
which jointly mimic the effects of waveform variation and electrode drift.
The results given in Figure \ref{drift}(b) show that CAE can compress and reconstruct not only spikes similar to the training data with high fidelity, but also unseen spikes exhibiting significant changes on amplitude and shape.

\begin{figure*}[t]
\center
\includegraphics[width=\linewidth]{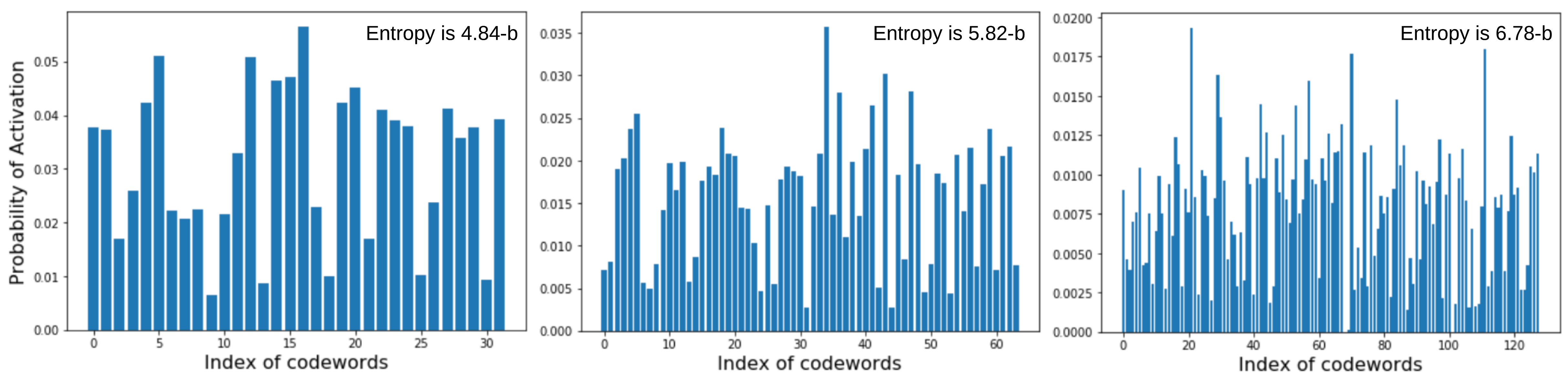}
\caption{{Activation patterns of the VQ codewords in CAE trained on \texttt{Wave\_Clus} dataset introduced in Section 3.1. From left to right, the numbers of codewords are 32, 64, and 128. The entropies of VQ codewords are 4.84-bit, 5.82-bit, and 6.78-bit.}}
\label{vq_pattern}
\vspace{-10pt}
\end{figure*}

{
We have done another experiment to evaluate the generalization capability of CAE by using different sequences for training and testing, including \texttt{C\_Difficult1} with noise levels 0.2 and 0.05, and \texttt{C\_Difficult2} with noise levels 0.2 and 0.05, all of which are from \texttt{Wave\_Clus} dataset.
\texttt{C\_Difficult1} and \texttt{C\_Difficult2} are synthesized using different spike templates.
For each sequence, the first 50\% of spikes are used for training and the rest 50\% for testing.}

{
The results given in Table \ref{cae_train} shows that CAE can generalize well against different spike templates and varied noise levels.
Specifically, (i) when spike templates are the same in the training and testing data, closer noise levels can lead to higher compression accuracies; 
(ii) when spike templates are different, higher noise levels can lead to better performance.
In the first case, VQ codewords learned from training data are closer to the spike components in testing data than noise due to the same templates.
In the second case, in the absence of common spike templates, CAE can learn more diverse features from noisier waveforms that better represent testing data in the latent space.
Performance in the first case is consistently better than in the second case.
The results that CAE can generalize well over different spike templates and background noises suggest the potential application of CAE in chronic wireless recording experiments to reliably compress neural signals without frequent re-training or parameter tuning.
}

\subsection{Effects of different numbers of VQ codewords}
{Another appealing feature of CAE is that it can uncover a low-dimensional space from spikes where features naturally conform to uniform distribution, which facilitates efficient and accurate vector quantization.}

{To understand this feature, we examine the activation patterns of VQ codewords.
As shown in Figure \ref{vq_pattern}, after trained on the \texttt{Wave\_Clus} dataset, the VQ codewords tend to be uniformly activated regardless of the number of codewords.
In other words, the entropy of VQ codewords is always close to $\text{log}_2K$.
The situation is similar on other datasets.
It should be noted that the uniformity is attained in the absence of any entropy regularization term in the loss function of CAE, which is normally required to enforce certain output distributions \cite{pmlr-v70-bojanowski17a}.
We hypothesize that CAE transforms spikes into a group of relatively invariant and uniformly distributed features inherent to spikes in the low-dimensional latent space, and VQ codewords converge to the grid-like spike features via nearest neighbor search.
Due to the uniform distribution of features, the convergence of VQ codewords can be fast, robust, and accurate.
}

\begin{figure*}[t]
\center
\includegraphics[width=\linewidth]{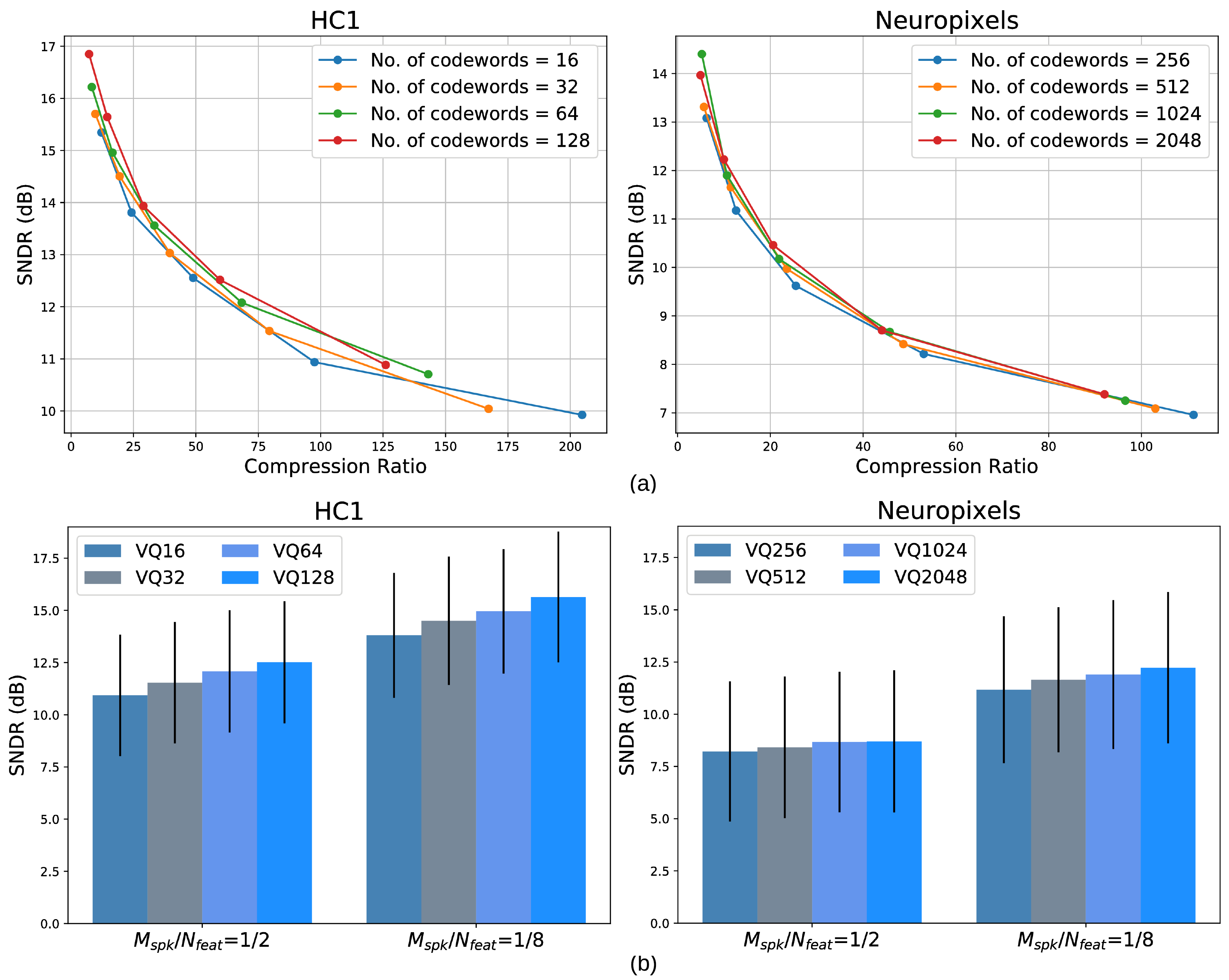}
\caption{Effects of different $M_{spk}/N_{feat}$ ratios and numbers of codewords on reconstruction accuracy. $K$ is varied from 16 to 128 for \texttt{HC1} and from 256 to 2048 for \texttt{Neuropixels}. (a) Rate-quality curves for both in-vivo dataset. (b) Comparison of accuracies at low and high $M_{spk}/N_{feat}$ ratios. Error bars representing standard variations of SNDR are labeled.}
\label{vq_invivo}
\vspace{-10pt}
\end{figure*}

{
Under this hypothesis, achieving higher accuracy is bottlenecked by the amount and quality of spike features output by the encoder, not VQ
codewords.}
To demonstrate this, we run CAE on the two in-vivo datasets (\texttt{HC1} and \texttt{Neuropixels}) with different numbers of codewords and plot their rate-quality curves.
As shown in Figure \ref{vq_invivo}(a), for each dataset the number of codewords is varied by up to 8$\times$ while the reconstruction accuracy changes by only $\sim$1dB.
{In comparison, increasing $N_{feat}$ by 4$\times$ can lead to 3--4dB improvement on SNDR, as shown in Figure \ref{vq_invivo}(b).}
This feature is useful for hardware implementation, since 
{the indexing logic that searches the nearest VQ codewords of encoder outputs can be simplified, thus reducing the processing delay and improving the throughput.}

{
Figure \ref{vq_invivo}(b) further shows that when spike morphologies are more complicated, the accuracy is more insensitive to the number of codewords.
Under this condition, it requires a larger $N_{feat}$ to allow for fine-grained sampling from the spike feature space; however, increasing $N_{feat}$ would decrease CR at the same time.
To maintain decent CRs, the inherent resolution of spike features is limited by the upper-bounded $N_{feat}$ and hence more codewords in this case will not be effective in improving accuracy.
}

\vspace{-5pt}
\subsection{Effects of preserving spatial proximity of spikes}
CAE is capable of extracting localized features from spikes recorded from channels that are geometrically closed to each other.
We have designed the following experiment to verify that CAE can leverage the geometric information of spikes to achieve higher accuracies at no cost: simply preserving their spatial proximity at the input to the network.

We picked 15 channels from the \texttt{Neuropixels} dataset along the longitudinal dimension of the probe with a spacing of 400$\mu$m.
This is to ensure that spikes detected from different channels are generated by different neurons and thus with independent waveform characteristics.
Two CAE models with the same configurations were created, where $K=512$, $M_{spk}=15$, and $M_{spk}/N_{feat}$ is set as 1, 1/2, 1/4, and 1/8.
The training and testing spikes for the two models are the same, except that \textbf{Model 1} was trained with spikes randomly shuffled along the channel dimension;
\textbf{Model 2} was trained with spikes preserving their spatial proximity, e.g., spikes detected from probe channel 1 are fed into CAE input port 1.

\begin{figure}[t]
\center
\includegraphics[width=\linewidth]{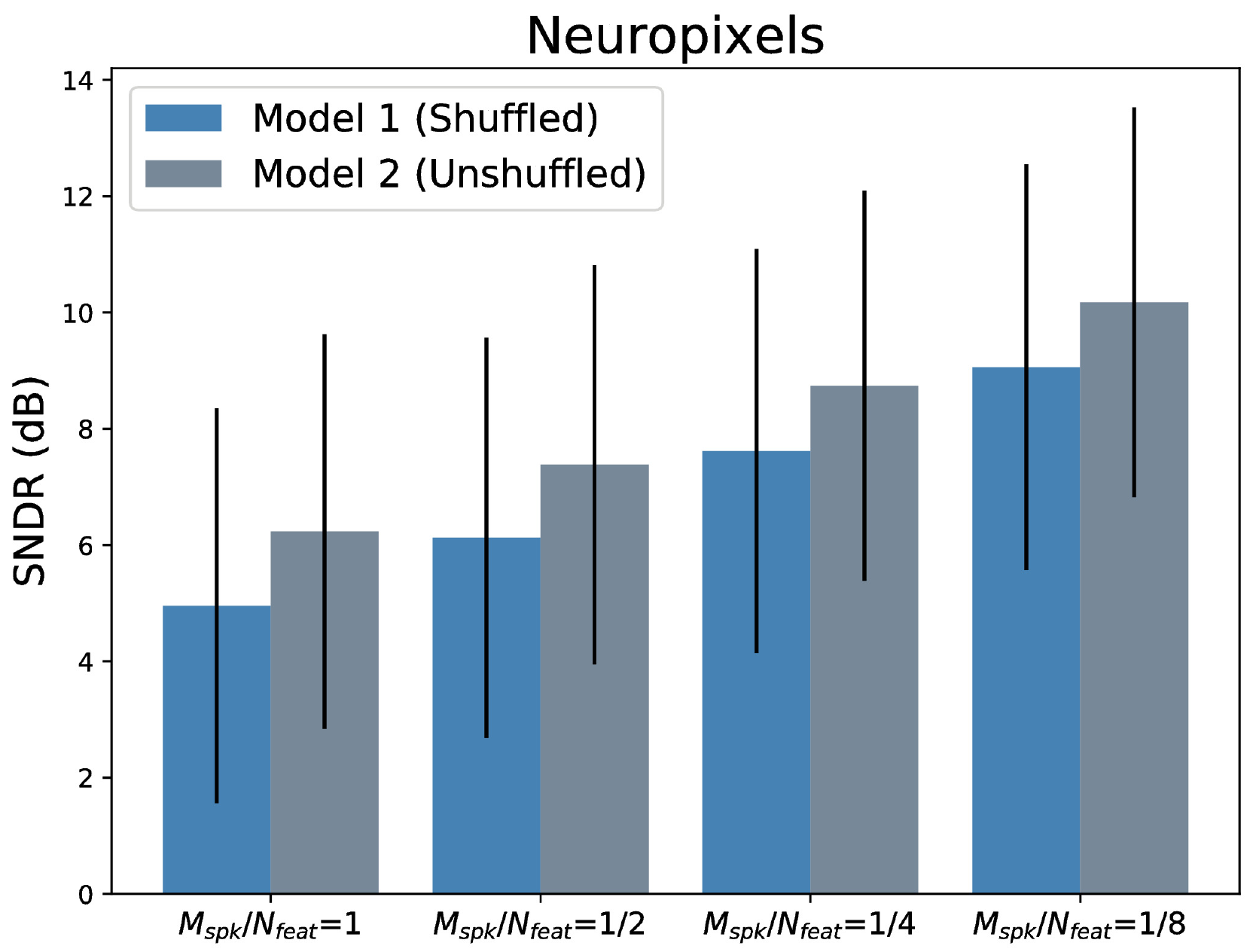}
\caption{Effects of shuffling spikes along channel dimension on compression accuracies. For both CAE models, $K=512$, $M_{spk}=15$, and $M_{spk}/N_{feat}$ is set as 1, 1/2, 1/4, and 1/8. Error bars represent standard deviations of SNDR.}
\label{shuffle}
\vspace{-10pt}
\end{figure}

\begin{figure*}[t]
\center
\includegraphics[width=\linewidth]{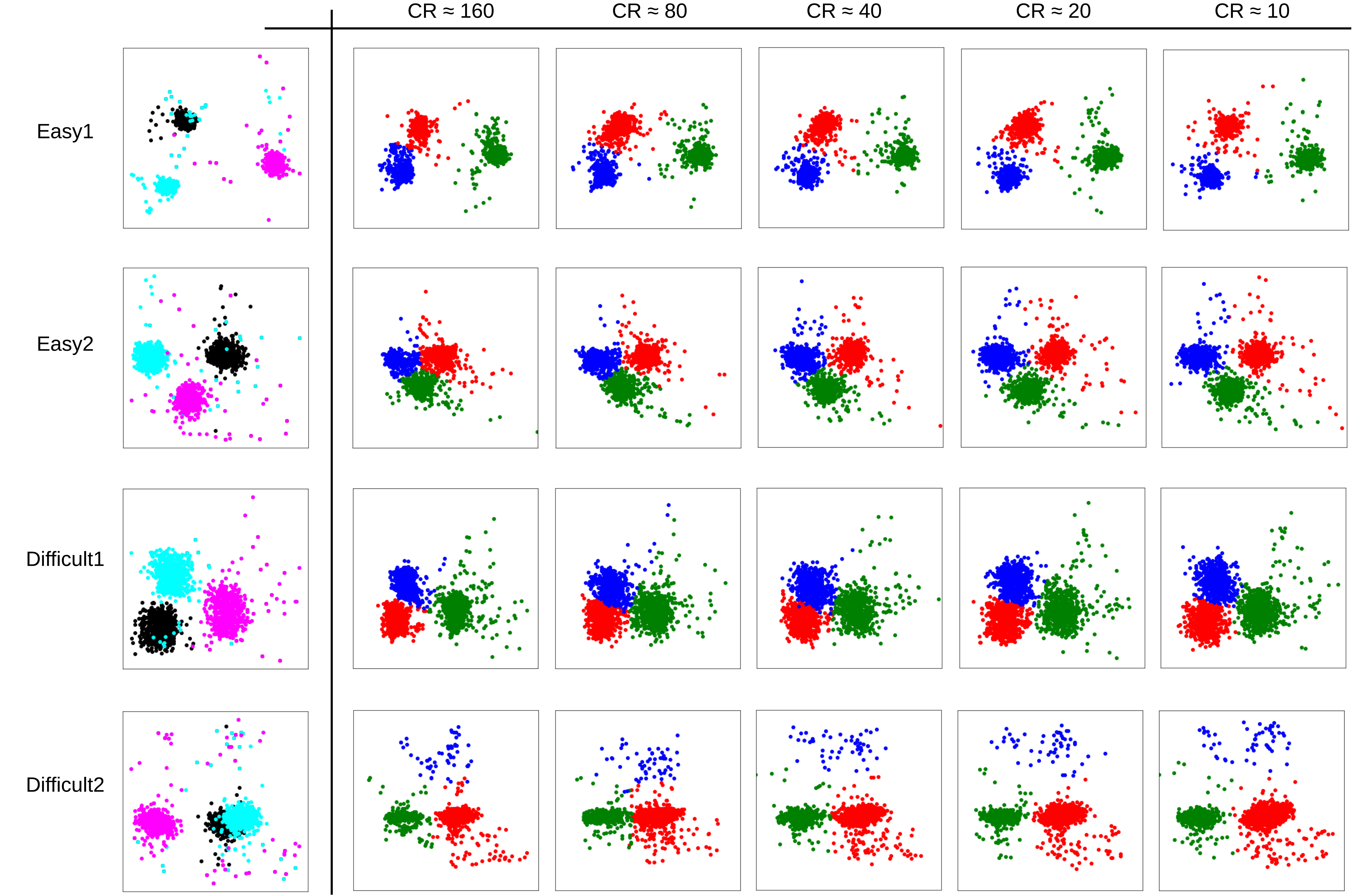}
\caption{{Visualization of spikes in 2-dimensional PCA space. From top to bottom, the four sequences are \texttt{C\_Easy1\_noise005}, \texttt{C\_Easy2\_noise005}, \texttt{C\_Difficult1\_noise005}, and \texttt{C\_Difficult2\_noise005}. Sequences with higher noise levels are not plotted because of poor separation in PCA feature space. Plots in the leftest column are the uncompressed spikes with ground truth labels. Plots in the second to the sixth columns are compressed spikes at different compression ratios classified with PCA + K-Means. In each row, compressed spikes are plotted in the same feature directions as the uncompressed spikes.}}
\label{scatter}
\vspace{-10pt}
\end{figure*}

\begin{figure*}[t]
\center
\includegraphics[width=\linewidth]{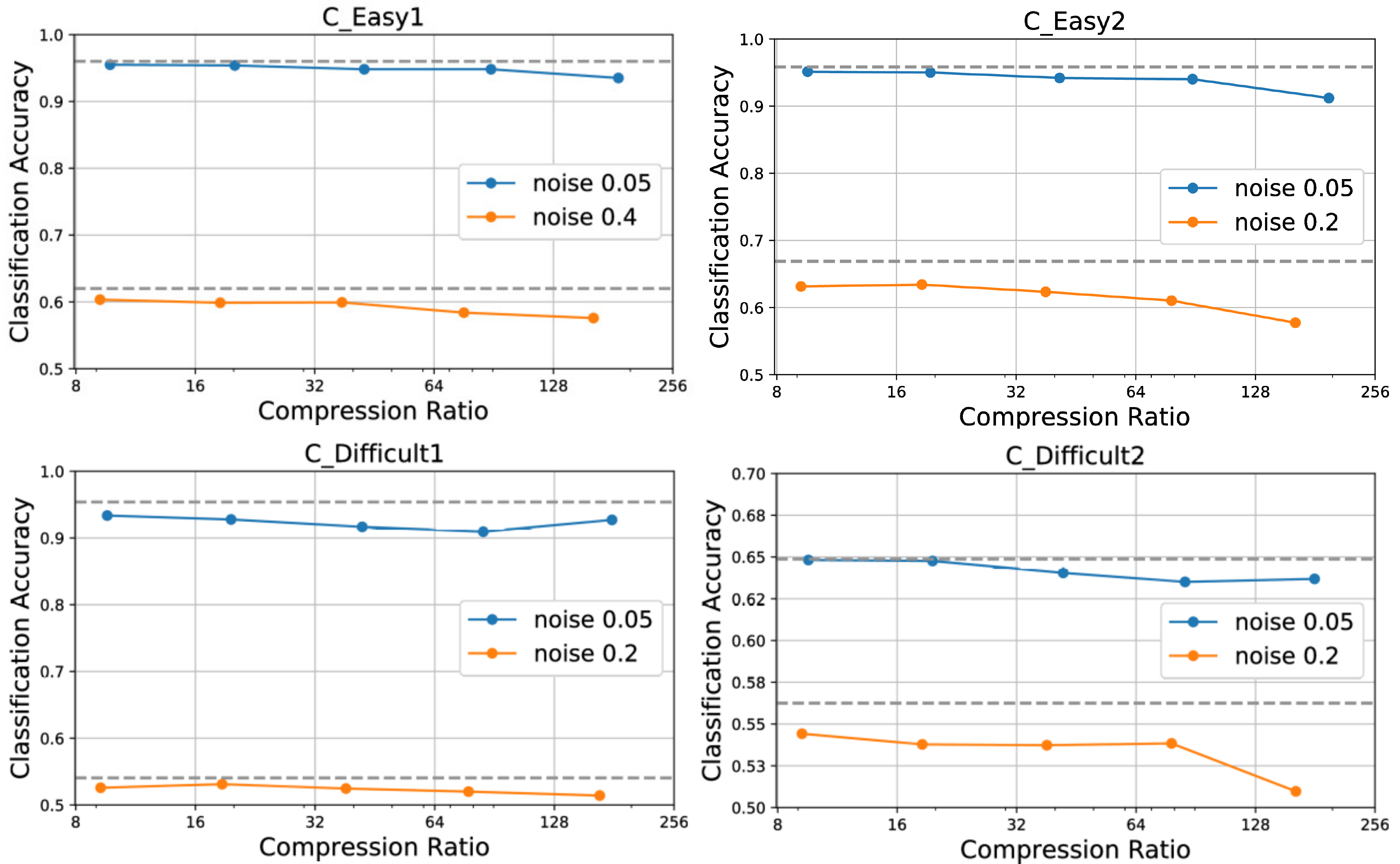}
\caption{{Evaluation of clustering (PCA + K-Means) performance before and after compression using CAE. For all datasets, $K=128$, $M_{spk}/N_{feat}$ is varied from 1/1 to 1/16. Feature dimension after PCA is 3. K-Means is run 500 times on each sequence after PCA with randomized centroid initializations. In each sub-figure, the two gray dashed lines represent the sorting accuracies before compression and correspond to the low and hight noise levels, respectively.}}
\label{clus}
\vspace{-10pt}
\end{figure*}

The testing performances of the two CAE models are given in Figure \ref{shuffle}.
With training spikes shuffled along the channel dimension, the spatial proximity was disrupted. 
The compression accuracy of \textbf{Model 1} is consistently poorer than that of \textbf{Model 2} by 1--2dB across different $M_{spk}/N_{feat}$ ratios; meanwhile, their CRs are comparable (not shown in Figure \ref{shuffle}).
Thus, preserving spatial proximity is important for CAE to extract localized features from multichannel spikes and achieve higher accuracies.
\vspace{-10pt}

{\subsection{Evaluation of clustering performance before and after compression}}
{An important analysis in neural signal processing is to obtain single-unit activity from raw recordings, a process commonly known as spike sorting that classifies spikes to their originating neurons \cite{Gibson-SPM-2012}. 
Hence it is necessary to evaluate the distortions on spike sorting accuracy introduced by compression.}
{We used the following datasets from \texttt{Wave\_Clus}: \texttt{C\_Easy1}, \texttt{C\_Easy2}, \texttt{C\_Difficult1}, and \texttt{C\_Difficult2}. 
From each dataset, we picked two sequences with the lowest and highest background noise levels, respectively.
Spikes were identified according to the ground truth timestamps.
In each sequence, the first 50\% of spikes were used to train the CAE model, and the rest were used for testing.
For each spike, the first 3 principal components are extracted as features using PCA.
We run K-Means 500 times on the principal components of spikes from each sequence with randomized centroid initializations to ensure the best classification result.
We repeated the spike sorting pipeline on each sequence compressed by CAE with different compression ratios, and compared the results with the ideal classification accuracies obtained from the uncompressed spikes.
}

In Figure \ref{scatter}, we visualize the testing spikes in the 2-dimensional PCA feature space at different CRs.
It shows that with smaller CRs (higher SNDR), compressed spikes tend to be more ``scattered'' and resemble the distribution patterns of uncompressed spikes.
The spike sorting results are given in Figure \ref{clus}.
In each sub-figure, the two gray dashed lines represent the sorting accuracies using uncompressed spikes with low and high noise levels, respectively.
With low noise levels, the drop of classification accuracy caused by CAE compression is less than 4\% for up to 178$\times$ CR;
with high noise levels, the drop of classification accuracy is slightly larger than with low noise levels, mostly less than 5\% except for \texttt{C\_Easy2} where the performance drops by 9\% at 161$\times$ CR.
In addition, on all sequences, the sorting accuracy stays almost unchanged with respect to exponentially increased CR (until CR is over 64$\times$).
The results suggest that CAE can allow for high CRs without noticeably compromising spike sorting performance.

\vspace{10pt}
\subsection{Evaluation of CAE performance in the presence of spike misalignment and overlapping}
For the purpose of reliable and accurate feature extraction, it is often required that spikes are aligned to the peaks or maximum slopes.
However, accurate spike alignment is difficult in low-SNR recordings due to sampling jitter combined with noise effects \cite{Gibson-SPM-2012}, and results in misaligned spikes.

{
One potential solution is to perform careful spike detection, which discerns ``clean'' spike shapes that can be well aligned from noisy backgrounds.
However, such operation is often supervised and time-consuming, and also computationally unrealistic for on-chip and real-time implementation.
To overcome this difficulty, spike compressor is expected to perform robustly against misalignment.
To examine this capability of CAE, we chose a low-SNR sequence from \texttt{Wave\_Clus} (\texttt{C\_Difficult2\_noise02}).
We added a small temporal jitter to each ground truth timestamp and extracted spikes.
The jitters were randomly sampled from a centered uniform distribution spanning a width from 2 to 10 points with an increment of 2.
CAE was trained using the jittered spikes and evaluated by attempting to reconstruct clean spikes from jittered spikes.}

\begin{figure}[t]
\center
\includegraphics[width=\linewidth]{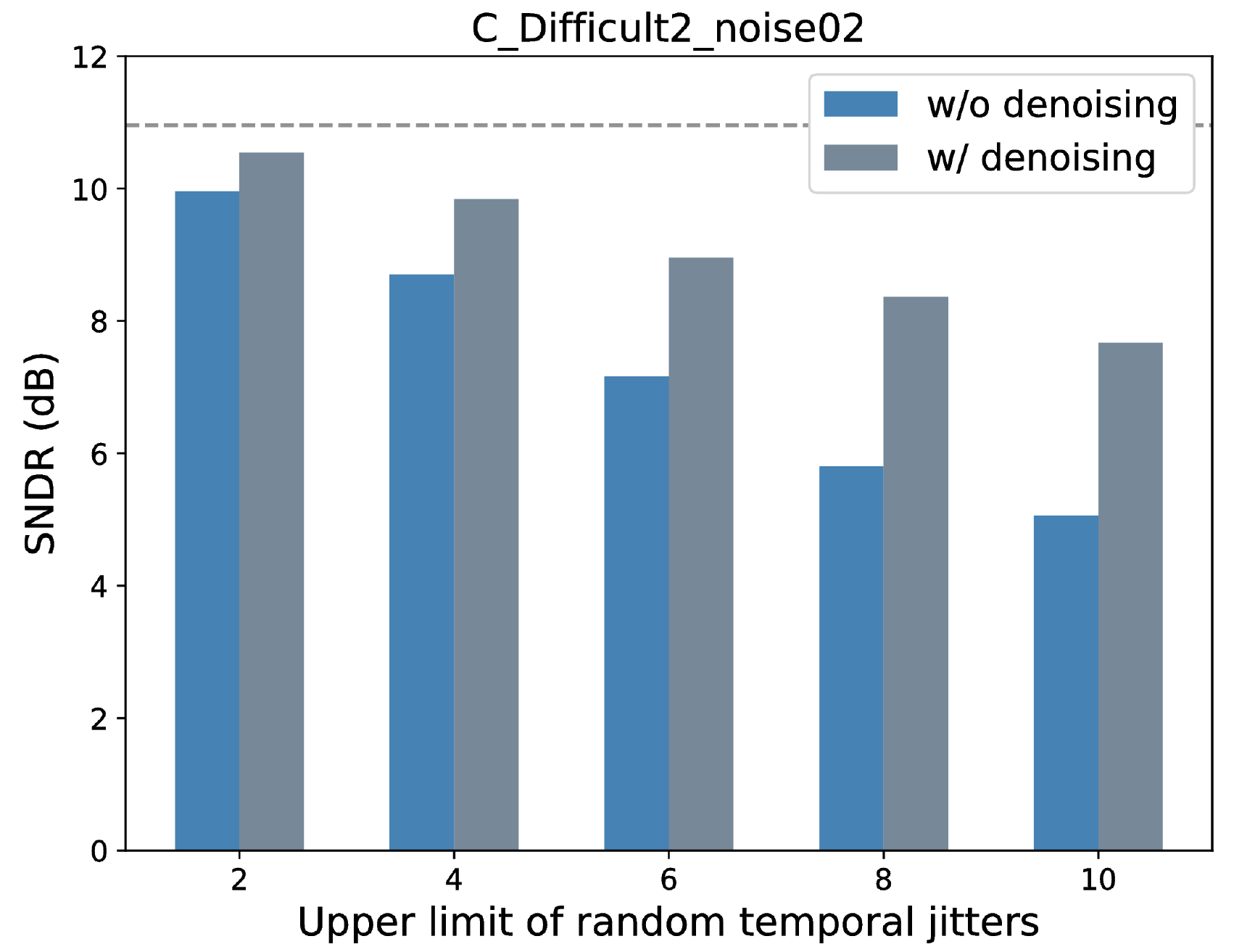}
\caption{{Performance of CAE against spike misalignment. Each spike is temporally jittered by up to 2, 4, 6, 8, 10 points. Compression accuracy is 10.95dB without misalignment (gray dashed line). Compression accuracies w/o and w/ denoising CAE are drawn in blue and gray bars, respectively.}}
\label{misalign}
\vspace{-10pt}
\end{figure}

{As shown in Figure \ref{misalign}, the compression accuracies decrease by 1--6 dB at different extents of misalignment (blue bars).
Here we present a technique to enhance the performance by configuring CAE as a denoising autoencoder without modifying its structure.
Referring to equation (3), instead of using the jittered spikes $\mathbf{x}$, we use the clean spikes $\mathbf{x}_{clean}$ that correspond to the jittered spikes to compute the loss as:
\begin{equation}
\mathcal{L}_{CAE} = d(\mathbf{x}_{clean}, \mathbf{\hat{x}}) + d(\mathbf{y}, \mathbf{\hat{y}}),
\end{equation}
where $\mathbf{y}$ and $\mathbf{\hat{y}}$ are still calculated from the jittered spikes $\mathbf{x}$.
The access to $\mathbf{x}_{clean}$ is feasible since CAE needs to be trained off-line, where training spikes can always be accurately aligned.
Optimized with the new loss, CAE is encouraged to learn reconstructing clean spikes from misaligned spikes -- an essentially denoising process, thus can perform substantially better on unseen spikes with similar misalignment (gray bars in Figure \ref{misalign}).
}

\begin{figure}[t]
\center
\includegraphics[width=\linewidth]{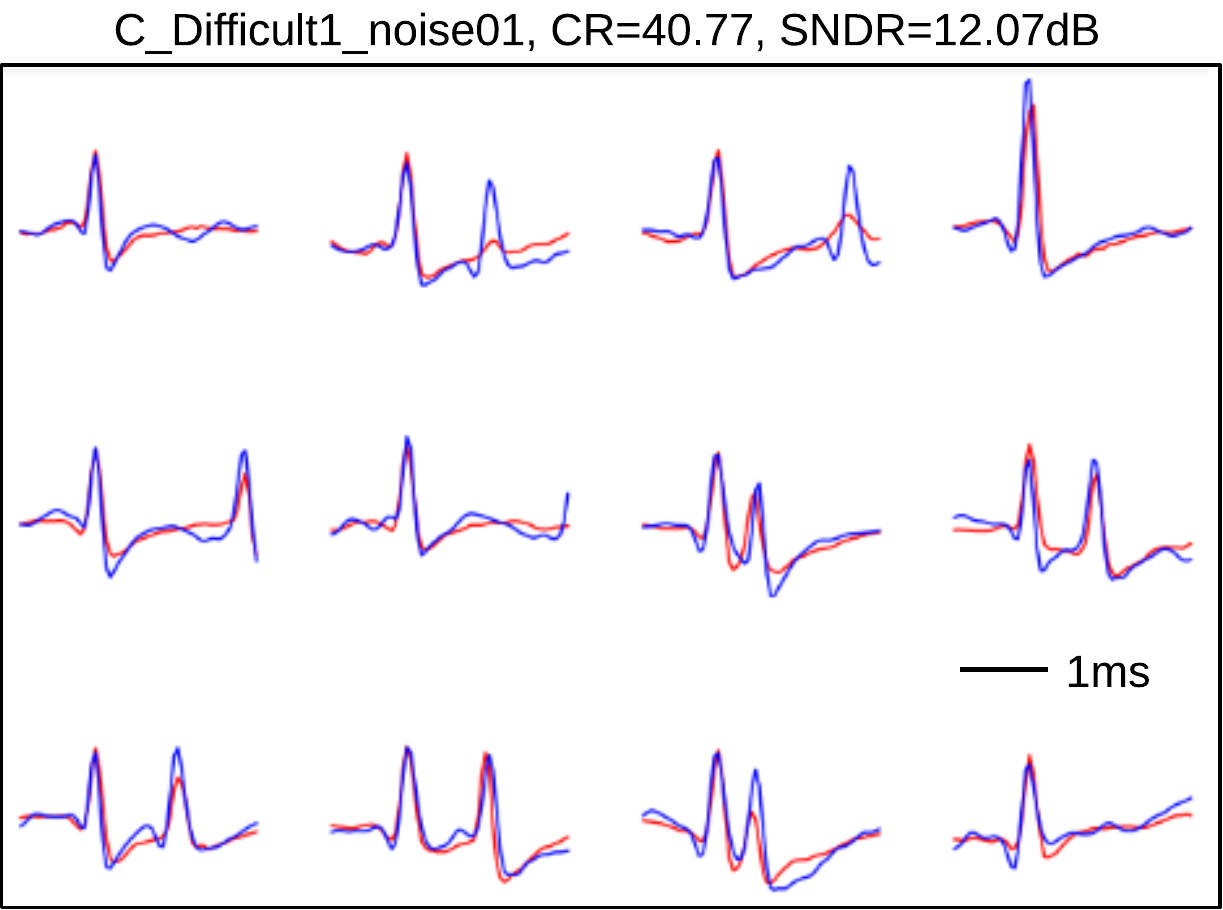}
\caption{{Performance of CAE against spike overlapping. Original and reconstructed spikes are drawn in blue and red, respectively. Configuration of CAE: $K=128$ and $M_{spk}/N_{feat}=1/4$.}}
\label{overlap}
\vspace{-10pt}
\end{figure}

{
Another issue that will hurt compression performance is spike overlapping, which can be frequent in high-density recordings, especially with high-rate spike activities.
Resolving overlapped spikes is a challenging task.
In the recent spike sorting pipelines (e.g. \texttt{KiloSort} \cite{pachitariu2016fast}), it is approached by comparing overlapped spikes with an exhaustive search of linear combinations of clean spike templates (typically two spikes with varied amplitudes and phases), and requires iterative processing that can only be afforded off-line.
In this work, our design goal is to make CAE capable of compressing both clean and overlapped spikes as accurately as possible, and leave the computationally expensive resolving overlapped spikes to off-line processing.
As shown in Figure \ref{overlap}, CAE shows a promising performance in representing both clean and overlapped spikes at a reasonably good CR.
}

\section{Discussion}

\subsection{Increasing channel dimension versus grouped convolution}
At the first glance, the elevation of channel dimension at the input of the CAE encoder from $M_{spk}$ to 256 that dramatically increases the total model parameters is in contradiction to the application of grouped convolution in all the \texttt{ResNeXt} modules that is used to reduce the amount of model parameters.

In computer vision applications, increasing channel dimension is a common practice and found useful to uncover effective features from images for the task of classification.
In spike compression, the proposed CAE bears a resemblance to those renowned models in computer vision, such as VGG-16 \cite{simonyan2014very}, AlexNet \cite{krizhevsky2012imagenet}, and ResNet-50 \cite{he2016deep}, in a way that relies on stacking CNN layers with increasing channel dimensions to extract representative features and VQ to essentially ``classifies'' the encoder outputs to the codewords.
Following the same intuition, we adopted a similar configuration in CAE and achieved good performance.

On the other hand, the amount of model parameters can be safely reduced without changing the channel dimension that is key for feature learning.
One of such solutions is grouped convolution.
It servers another purpose in our application, which is to preserve localized features from recording channels that are geometrically proximate, as detailed in Section 3.4.

\subsection{Hand-crafted features versus CNN features}
A key step to achieve good performance for all types of compression methods is to find a feature space where a more compact and effective representation of neural signals can be identified to facilitate compression.

In this paper, the proposed CAE model was compared with four commonly used methods for data compression: PCA, DWT, DCT, and compressive sensing, each of which represents a type of methods operating in a distinctive feature space.
PCA converts signals into an orthogonal space with the directions of axes capturing the data variance in the descending order (\textbf{orthogonal domain});
DWT ``quantizes'' signals with shifted and scaled versions of mother wavelet (\textbf{wavelet domain});
DCT converts signals into the frequency domain by expressing them as sums of cosine functions oscillating at different frequencies (\textbf{frequency domain});
compressive sensing exploits the signal sparsity by encoding them with randomized projections and reconstructing signals using fewer samples required by Shannon-Nyquist sampling theorem (\textbf{sparse domain}).
In general, the application of the above transformations on neural signals is heuristic and not optimized for reconstruction accuracy nor compression ratio in a principled way.
In comparison, the hierarchical features extracted by CAE in the latent space are trained to optimize the reconstruction quality thus accuracy is guaranteed;
meanwhile, compression is gained from the dimensionality reduction achieved by the encoder and the bottleneck structure of the network (low-dimensional latent space) as well as the entropy coding based on VQ results.
This is analogous to the comparison between scale invariant feature transform (SIFT) and CNN as feature detectors in image processing, in which SIFT is hard coded, low-level gradient-based feature whereas CNN features are obtained through hierarchical layer-wise representation learning.
Furthermore, CAE features can be trained end-to-end, which avoids any heuristic constructions that may limit performance.

\subsection{Feasibility of on-chip hardware implementation and estimation of online performance}
Deployment of deep learning models onto hardware platforms with limited resources and constraints of power/heat dissipation is challenging due to the excessive amount of model parameters and incurred computations.
For example, \texttt{ResNet-152}, the first deep learning model that won the ImageNet classification challenge by surpassing human-level accuracy, contains 60 million weights and requires 11.3G multiply-accumulates (MAC) to process one image \cite{sze2017efficient}.
Consequently, the model size of \texttt{ResNet-152} is over 200MByte.
It is impractical to implement compression models with similar sizes as application-specific integrated circuit (ASIC) chips.
Hence it is crucial to take into serious consideration the complexity of compression models for on-chip integration with analog frontend recording circuitry.

We first examine the model size of CAE.
CAE contains 8 convolutional layers in the encoder network interleaved with pooling and normalization layers which require none or trivial amount of parameters.
Table \ref{enc_model_size} and \ref{dec_model_size} give detailed information of a CAE model trained on \texttt{Neuropixels} dataset for the on-chip and off-chip parts, respectively.
The output size of each layer follows the format of \{\texttt{batch size}, \texttt{channel dimension}, \texttt{feature dimension}\}.
The parameters of the encoder and VQ are counted together as they are to be on-chip implemented.
The total amount of parameters is 812K, which is a minor fraction of that of \texttt{ResNet-152};
furthermore, the encoder (including VQ) is over 44$\times$ smaller than the decoder, thanks to the grouped convolution technique employed for the \texttt{ResNeXt} module, resulting in fewer than 18K parameters.
Assuming an 8-bit weight precision (which has been used successfully in several commercial products such as Tensor Processing Unit), it would require 18KByte memory to store the CAE on-chip.
Taking \texttt{Eyeriss} \cite{chen2017Eyeriss} as a reference design (one of the state-of-the-art deep learning chips), it has 181.5KByte on-chip SRAM and 108KByte global buffers, both of which are sufficient to load the on-chip part of CAE.

Next we examine the computational complexity and power efficiency of the on-chip part of the CAE model reported in Table \ref{enc_model_size}.
On average, it takes 79.25K MACs to process one spike for on-chip computation.
To estimate the power efficiency, again we refer to \texttt{Eyeriss} as a reference design, which has an energy efficiency of 83.1GMACs per Watt \cite{chen2017Eyeriss}.
Therefore, the on-chip part of CAE would consume 0.95$\mu$W to process one spike if implemented on \texttt{Eyeriss}.
Assuming an average firing rate of 20Hz per channel, the power consumption of spike compression using CAE would be 19$\mu$W/channel, which is comparable to that of analog recording circuitry (10--50$\mu$W/channel \cite{xu2014frequency}).

\begin{table}[t]
\centering
\caption{Model parameters of encoder and VQ}
\label{enc_model_size}
\begin{threeparttable}
\begin{tabular}{|m{3.15cm}<{\centering}|c|c|}
\hline
Layers              		& Output Size       	& Parameters      \\ \hline
Convolution         		& N\tnote{a}, 256, 48  	& 1024            \\ \hline
Normalization       		& N, 256, 48      		& 512             \\ \hline
\texttt{ResNeXt} (3$\times$conv.) 	& N, 256, 48      		& 4608            \\ \hline
Downsampling        		& N, 256, 24      		& 0               \\ \hline
\texttt{ResNeXt} (3$\times$conv.) 	& N, 256, 24      		& 4608            \\ \hline
Downsampling        		& N, 256, 12      		& 0               \\ \hline
Convolution         		& N, 16, 12       		& 4096            \\ \hline
Normalization       		& N, 16, 12       		& 32              \\ \hline
Vector Quantization (256 codewords) & N$\times$16, 12        & 3072            \\ \hline
\textbf{Total}      & \multicolumn{2}{c|}{\textbf{17952}} \\ \hline
\end{tabular}
\begin{tablenotes}
\footnotesize
\item[a] N denotes batch size in Table \ref{enc_model_size} and \ref{dec_model_size}.
\end{tablenotes}
\end{threeparttable}
\vspace{-10pt}
\end{table}

\begin{table}[t]
\centering
\caption{Model parameters of decoder}
\label{dec_model_size}
\begin{tabular}{|c|c|c|}
\hline
Layers              		& Output Size       & Parameters      \\ \hline
Deconvolution         		& N, 256, 12      	& 4096            \\ \hline
Normalization       		& N, 256, 12      	& 512             \\ \hline
Upsampling        			& N, 256, 24      	& 0               \\ \hline
\texttt{ResNet} (2$\times$deconv.) 	& N, 256, 24      	& 394240          \\ \hline
Upsampling        			& N, 256, 48      	& 0               \\ \hline
\texttt{ResNet} (2$\times$deconv.) 	& N, 256, 48      	& 394240          \\ \hline
Deconvolution         		& N, 4, 48       	& 1028            \\ \hline
\textbf{Total}      		& \multicolumn{2}{c|}{\textbf{794116}} \\ \hline
\end{tabular}
\vspace{-10pt}
\end{table}

Regarding the processing speed of the CAE model, \texttt{Eyeriss} has a throughput of at least 16.8GMACs.
Referring to the requirement of 79.25KMACs/spike derived earlier, the on-chip CAE model has a theoretical peak throughput of 0.22M spikes.
However, this astonishing processing capability cannot be achieved, because the power density of invasive neural implants that conduct brain signal sensing, processing, and transmission must adhere to rigid regulations, that is smaller than 400$\mu$W/mm$^2$, to prevent from damaging brain tissues caused by increased temperatures as a result of heat dissipation \cite{wolf2008thermal}.
The power density of \texttt{Eyeriss} is 22.67mW/mm$^2$ at a throughput of 23.1GMACs.
Constraining the power density to 400$\mu$W/mm$^2$, the highest throughput is 0.4GMACs, which translates to processing around 5000 spikes simultaneously.
As neuronal firing is in general sparse and concurrent firing of multiple nearby neurons is infrequent, the 5000 spikes throughput should be able to support thousands of recording channels simultaneously.

\section{Conclusion}
In this work, we propose CAE, a DNN-based spike compression model to significantly reduce the data rate of spikes in large-scale neural recording experiments.
Compared with existing methods, the proposed approach can raise the CR to 20--500$\times$ while provides comparable or better signal qualities.
Through extensive experiments, we have shown several advantageous features of CAE:
{(i) CAE can extract representative features from spikes, which are robust to non-stationarities of neural activities (waveform variation and electrode drift) and recording imperfections (spike misalignment and overlapping);}
(ii) Thanks to the VQ implemented in the latent space, the reconstruction accuracy of CAE is insensitive to the number of codewords for quantization, leading to improved processing throughput and simplified indexing logic;
(iii) CAE is capable of leveraging the spatial proximity of spikes from multiple channels to improve compression performance;
{(iv) CAE allows for high compression ratios without noticeably compromising spike sorting accuracy;}
(v) CAE features an asymmetric model structure, in which the encoder can be designed in a way that requires much less hardware resources than the decoder without undermining feature extraction capability, thus making CAE very suitable for hardware-efficient deployment into implantable neural recording systems.
We also provided quantitative evaluation of implementing CAE on a recent state-of-the-art deep learning acceleration chip \texttt{Eyeriss}, and demonstrated the potential to support thousands of recording channels simultaneously for spike compression.

\section*{Acknowledgment}
This work was supported by a DARPA grant HR0011-17-2-0060 and the University of Minnesota internal funding through MnDRIVE.
We would like to thank Dr. Wing-kin Tam for useful discussion on designing DNN models; Dr. Nick Steinmetz and the CortexLab at the University College London for providing the \texttt{Neuropixels} dataset; the NVIDIA corporation for generous donation of GPU resources.
We also would like to thank the reviewers for their useful comments to improve the quality of this work.

\begin{figure*}[t]
\center
\includegraphics[width=.9\linewidth]{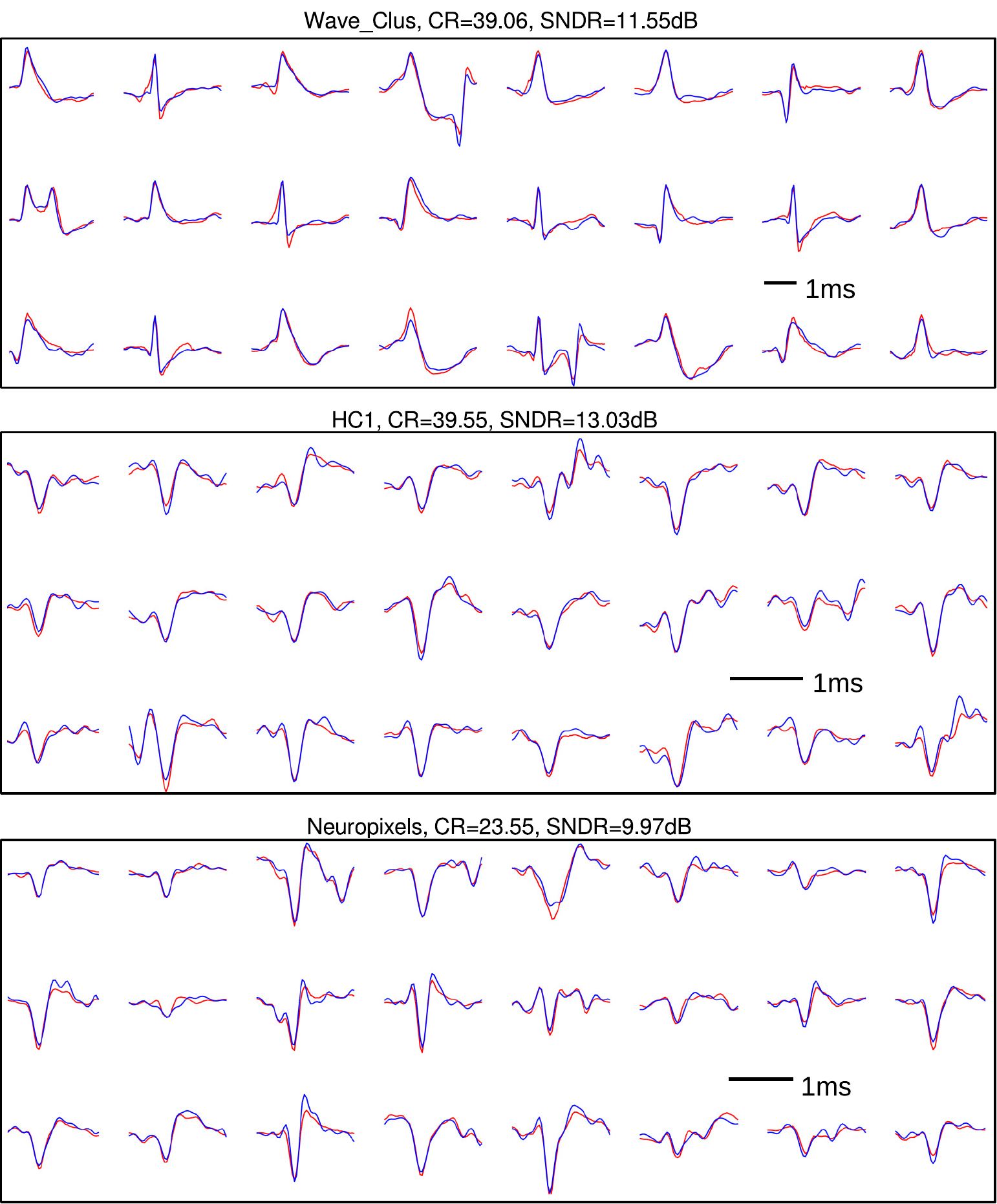}
\caption{Reconstructed spikes using CAE. 24 spikes with various shapes are chosen and shown for each dataset. Original and reconstructed spikes are drawn in blue and red colors, respectively. Configuration of CAE: $M_{spk}/N_{feat}=1/4$ for all datasets; $K=128, 32, 512$ for \texttt{Wave\_Clus}, \texttt{HC1}, and \texttt{Neuropixels} datasets, respectively. The reported CR and SNDR on top of each sub-figure are calculated over the entire testing part of corresponding dataset.}
\label{recon_spks}
\vspace{-10pt}
\end{figure*}

\begin{figure*}[t]
\center
\includegraphics[width=.9\linewidth]{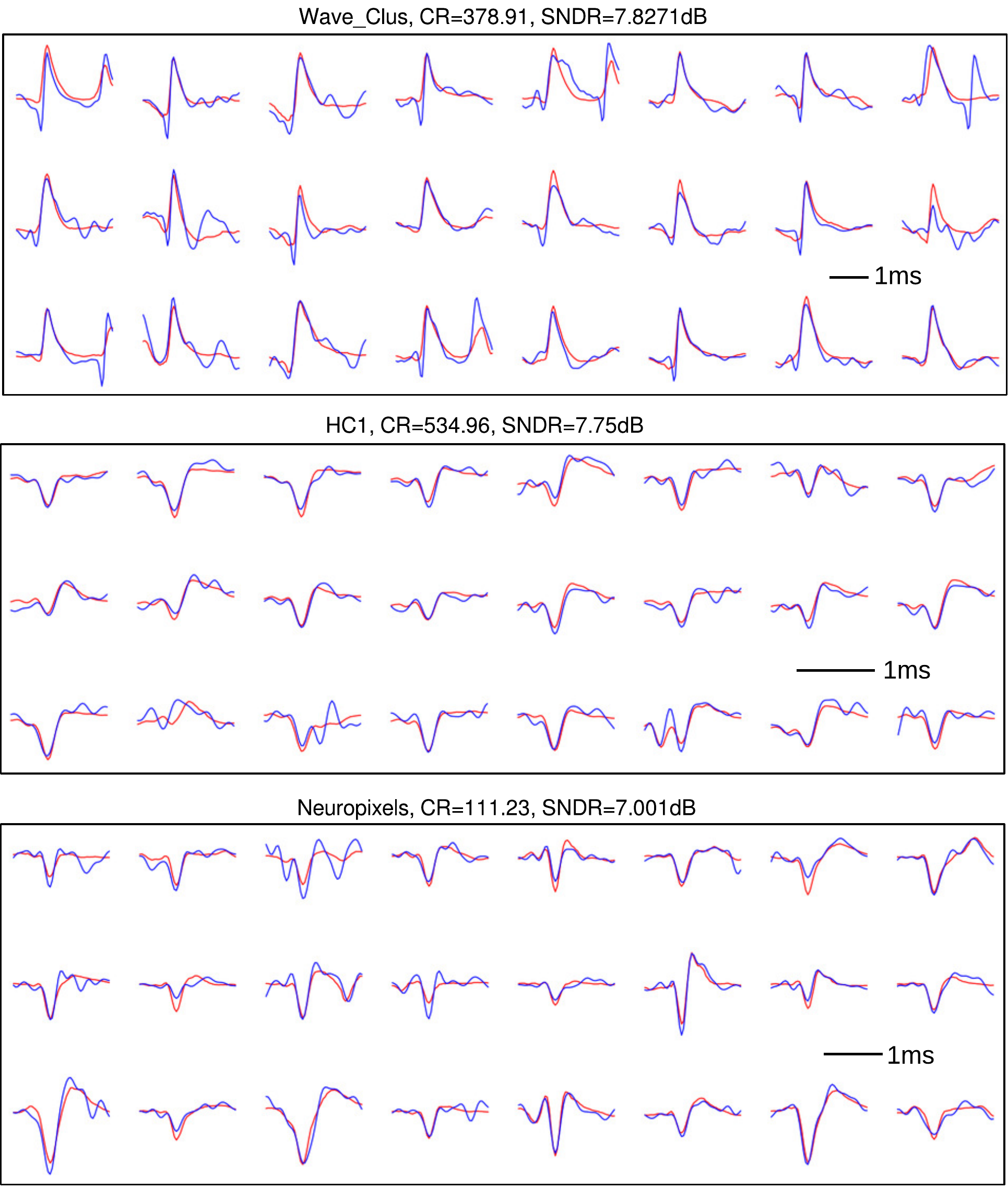}
\caption{{Reconstructed spikes using CAE. 24 spikes with various shapes are chosen and shown for each dataset. Original and reconstructed spikes are drawn in blue and red colors, respectively. Configuration of CAE: $M_{spk}/N_{feat}=2$ for all datasets; $K=128, 16, 256$ for \texttt{Wave\_Clus}, \texttt{HC1}, and \texttt{Neuropixels} datasets, respectively. The reported CR and SNDR on top of each sub-figure are calculated over the entire testing part of corresponding dataset.}}
\label{spk_high_cr}
\vspace{-10pt}
\end{figure*}

\appendix
{
\section{Examples of reconstructed spikes at different compression ratios}
Figure \ref{recon_spks} and \ref{spk_high_cr} show examples of reconstructed spikes using CAE at different CRs.}

\section*{Reference}
\begin{thebibliography}{10}

\bibitem{insel2013nih}
Thomas~R Insel, Story~C Landis, and Francis~S Collins.
\newblock The nih brain initiative.
\newblock {\em Science}, 340(6133):687--688, 2013.

\bibitem{darpa2016nesd}
Neural engineering system design (nesd).
\newblock \url{https://www.darpa.mil/program/neural-engineering-system-design}.

\bibitem{berenyi2014large}
Antal Ber{\'e}nyi, Zolt{\'a}n Somogyv{\'a}ri, Anett~J Nagy, Lisa Roux, John~D
  Long, Shigeyoshi Fujisawa, Eran Stark, Anthony Leonardo, Timothy~D Harris,
  and Gy{\"o}rgy Buzs{\'a}ki.
\newblock Large-scale, high-density (up to 512 channels) recording of local
  circuits in behaving animals.
\newblock {\em Journal of neurophysiology}, 111(5):1132--1149, 2014.

\bibitem{yang2016embc}
Zhi Yang, Jian Xu, Anh~Tuan Nguyen, Tong Wu, Wenfeng Zhao, and Teris Tam.
\newblock Neuronix enables continuous, simultaneous neural recording and
  electrical microstimulation.
\newblock In {\em Engineering in Medicine and Biology Society (EMBC), 2016 38th
  Annual International Conference of the IEEE}, pages 1--4. IEEE, 2016.

\bibitem{jun2017fully}
James~J Jun, Nicholas~A Steinmetz, Joshua~H Siegle, Daniel~J Denman, Marius
  Bauza, Brian Barbarits, Albert~K Lee, Costas~A Anastassiou, Alexandru Andrei,
  {\c{C}}a{\u{g}}atay Ayd{\i}n, et~al.
\newblock Fully integrated silicon probes for high-density recording of neural
  activity.
\newblock {\em Nature}, 551(7679):232, 2017.

\bibitem{tsai2015high}
David Tsai, Esha John, Tarun Chari, Rafael Yuste, and Kenneth Shepard.
\newblock High-channel-count, high-density microelectrode array for closed-loop
  investigation of neuronal networks.
\newblock In {\em Engineering in Medicine and Biology Society (EMBC), 2015 37th
  Annual International Conference of the IEEE}, pages 7510--7513. IEEE, 2015.

\bibitem{stevenson2011advances}
Ian~H Stevenson and Konrad~P Kording.
\newblock How advances in neural recording affect data analysis.
\newblock {\em Nature neuroscience}, 14(2):139--142, 2011.

\bibitem{Karkare-JSSC-2011}
V.~Karkare, S.~Gibson, and D.~Markovi\'{c}.
\newblock A 130-$\mu${W}, 64-{C}hannel {N}eural {S}pike-{S}orting {DSP} {C}hip.
\newblock {\em IEEE J. Solid-State Circuits}, 46:1214--1222, 2011.

\bibitem{7055372}
T.~Wu, J.~Xu, Y.~Lian, A.~Khalili, A.~Rastegarnia, C.~Guan, and Z.~Yang.
\newblock A 16-channel nonparametric spike detection asic based on ec-pc
  decomposition.
\newblock {\em IEEE Transactions on Biomedical Circuits and Systems},
  10(1):3--17, Feb 2016.

\bibitem{wu2015nonparametric}
Tong Wu, Jian Xu, and Zhi Yang.
\newblock A nonparametric neural signal processor for online data compression
  and power management.
\newblock In {\em Biomedical Circuits and Systems Conference (BioCAS), 2015
  IEEE}, pages 1--4. IEEE, 2015.

\bibitem{Gosselin-TBCAS-2009}
B.~Gosselin, A.~E. Ayoub, J.~Roy, M.~Sawan, F.~Lepore, A.~Chaudhuri, and
  D.~Guitton.
\newblock A {M}ixed-{S}ignal {M}ultichip {N}eural {R}ecording {I}nterface
  {W}ith {B}andwidth {R}eduction.
\newblock {\em IEEE Trans. Biomed. Circuits Syst.}, 3:129--141, 2009.

\bibitem{balle2016end}
Johannes Ball{\'e}, Valero Laparra, and Eero~P Simoncelli.
\newblock End-to-end optimized image compression.
\newblock {\em arXiv preprint arXiv:1611.01704}, 2016.

\bibitem{wu2017streaming}
Tong Wu, Wenfeng Zhao, Hongsun Guo, Hubert~H Lim, and Zhi Yang.
\newblock A streaming pca vlsi chip for neural data compression.
\newblock {\em IEEE transactions on biomedical circuits and systems},
  11(6):1290--1302, 2017.

\bibitem{chae2008128}
Moosung Chae, Wentai Liu, Zhi Yang, Tungchien Chen, Jungsuk Kim, Mohanasankar
  Sivaprakasam, and Mehmet Yuce.
\newblock A 128-channel 6mw wireless neural recording ic with on-the-fly spike
  sorting and uwb tansmitter.
\newblock In {\em Solid-State Circuits Conference, 2008. ISSCC 2008. Digest of
  Technical Papers. IEEE International}, pages 146--603. IEEE, 2008.

\bibitem{shinde2011comparison}
Amita~A Shinde and Pramod Kanjalkar.
\newblock The comparison of different transform based methods for ecg data
  compression.
\newblock In {\em Signal Processing, Communication, Computing and Networking
  Technologies (ICSCCN), 2011 International Conference on}, pages 332--335.
  IEEE, 2011.

\bibitem{zordan2014performance}
Davide Zordan, Borja Martinez, Ignasi Vilajosana, and Michele Rossi.
\newblock On the performance of lossy compression schemes for energy
  constrained sensor networking.
\newblock {\em ACM Transactions on Sensor Networks (TOSN)}, 11(1):15, 2014.

\bibitem{shaeri2015method}
Mohammad~Ali Shaeri and Amir~M Sodagar.
\newblock A method for compression of intra-cortically-recorded neural signals
  dedicated to implantable brain--machine interfaces.
\newblock {\em Neural Systems and Rehabilitation Engineering, IEEE Transactions
  on}, 23(3):485--497, 2015.

\bibitem{hosseini2014data}
Hossein Hosseini-Nejad, Abumoslem Jannesari, and Amir~M Sodagar.
\newblock Data compression in brain-machine/computer interfaces based on the
  walsh--hadamard transform.
\newblock {\em Biomedical Circuits and Systems, IEEE Transactions on},
  8(1):129--137, 2014.

\bibitem{zhao2018chip}
Wenfeng Zhao, Biao Sun, Tong Wu, and Zhi Yang.
\newblock On-chip neural data compression based on compressed sensing with
  sparse sensing matrices.
\newblock {\em IEEE Transactions on Biomedical Circuits and Systems}, 2018.

\bibitem{zhang2015closed}
Jie Zhang, Srinjoy Mitra, Yuanming Suo, Andrew Cheng, Tao Xiong, Frederic
  Michon, Marleen Welkenhuysen, Fabian Kloosterman, Peter~S Chin, Steven Hsiao,
  et~al.
\newblock A closed-loop compressive-sensing-based neural recording system.
\newblock {\em Journal of neural engineering}, 12(3):036005, 2015.

\bibitem{zhang2016communication}
Jie Zhang, Kerron Duncan, Yuanming Suo, Tao Xiong, Srinjoy Mitra, Trac~Duy
  Tran, and Ralph Etienne-Cummings.
\newblock Communication channel analysis and real time compressed sensing for
  high density neural recording devices.
\newblock {\em IEEE Transactions on Circuits and Systems I: Regular Papers},
  63(5):599--608, 2016.

\bibitem{sun2017training}
Biao Sun, Wenfeng Zhao, and Xinshan Zhu.
\newblock Training-free compressed sensing for wireless neural recording using
  analysis model and group weighted-minimization.
\newblock {\em Journal of neural engineering}, 14(3):036018, 2017.

\bibitem{craciun2011wireless}
Stefan Craciun, David Cheney, Karl Gugel, Justin~C Sanchez, and Jose~C
  Principe.
\newblock Wireless transmission of neural signals using entropy and mutual
  information compression.
\newblock {\em IEEE Transactions on Neural Systems and Rehabilitation
  Engineering}, 19(1):35--44, 2011.

\bibitem{bengio2009learning}
Yoshua Bengio et~al.
\newblock Learning deep architectures for ai.
\newblock {\em Foundations and trends{\textregistered} in Machine Learning},
  2(1):1--127, 2009.

\bibitem{Quiroga-NeuralComput-2004}
R.~Q. Quiroga, Z.~Nadasdy, and Y.~Ben-Shaul.
\newblock Unsupervised spike detection and sorting with wavelets and
  superparamagnetic clustering.
\newblock {\em Neural Comput.}, 16:1661--1687, 2004.

\bibitem{theis2017lossy}
Lucas Theis, Wenzhe Shi, Andrew Cunningham, and Ferenc Husz{\'a}r.
\newblock Lossy image compression with compressive autoencoders.
\newblock {\em arXiv preprint arXiv:1703.00395}, 2017.

\bibitem{NIPS2017_7210}
Aaron van~den Oord, Oriol Vinyals, and koray kavukcuoglu.
\newblock Neural discrete representation learning.
\newblock In I.~Guyon, U.~V. Luxburg, S.~Bengio, H.~Wallach, R.~Fergus,
  S.~Vishwanathan, and R.~Garnett, editors, {\em Advances in Neural Information
  Processing Systems 30}, pages 6306--6315. Curran Associates, Inc., 2017.

\bibitem{hu2017squeeze}
Jie Hu, Li~Shen, and Gang Sun.
\newblock Squeeze-and-excitation networks.
\newblock {\em arXiv preprint arXiv:1709.01507}, 2017.

\bibitem{cohen2018boosting}
Nadav Cohen, Ronen Tamari, and Amnon Shashua.
\newblock Boosting dilated convolutional networks with mixed tensor
  decompositions.
\newblock In {\em International Conference on Learning Representations}, 2018.

\bibitem{he2016deep}
Kaiming He, Xiangyu Zhang, Shaoqing Ren, and Jian Sun.
\newblock Deep residual learning for image recognition.
\newblock In {\em Proceedings of the IEEE conference on computer vision and
  pattern recognition}, pages 770--778, 2016.

\bibitem{veit2016residual}
Andreas Veit, Michael~J Wilber, and Serge Belongie.
\newblock Residual networks behave like ensembles of relatively shallow
  networks.
\newblock In {\em Advances in Neural Information Processing Systems}, pages
  550--558, 2016.

\bibitem{krizhevsky2012imagenet}
Alex Krizhevsky, Ilya Sutskever, and Geoffrey~E Hinton.
\newblock Imagenet classification with deep convolutional neural networks.
\newblock In {\em Advances in neural information processing systems}, pages
  1097--1105, 2012.

\bibitem{xie2017aggregated}
Saining Xie, Ross Girshick, Piotr Doll{\'a}r, Zhuowen Tu, and Kaiming He.
\newblock Aggregated residual transformations for deep neural networks.
\newblock In {\em Computer Vision and Pattern Recognition (CVPR), 2017 IEEE
  Conference on}, pages 5987--5995. IEEE, 2017.

\bibitem{gray2012source}
Robert~M Gray.
\newblock {\em Source coding theory}, volume~83.
\newblock Springer Science \& Business Media, 2012.

\bibitem{paszke2017automatic}
Adam Paszke, Sam Gross, Soumith Chintala, Gregory Chanan, Edward Yang, Zachary
  DeVito, Zeming Lin, Alban Desmaison, Luca Antiga, and Adam Lerer.
\newblock Automatic differentiation in pytorch.
\newblock In {\em NIPS-W}, 2017.

\bibitem{kingma2014adam}
Diederik~P Kingma and Jimmy Ba.
\newblock Adam: A method for stochastic optimization.
\newblock {\em arXiv preprint arXiv:1412.6980}, 2014.

\bibitem{harris2000accuracy}
Kenneth~D Harris, Darrell~A Henze, Jozsef Csicsvari, Hajime Hirase, and
  Gy{\"o}rgy Buzs{\'a}ki.
\newblock Accuracy of tetrode spike separation as determined by simultaneous
  intracellular and extracellular measurements.
\newblock {\em Journal of neurophysiology}, 84(1):401--414, 2000.

\bibitem{Henze-JNeuronphy-2000}
Henze~D A, Borhegyi Z, Csicsvari J, Mamiya A, Harris~K D, and Buzsaki G.
\newblock {I}ntracellular {F}eatures {P}redicted by {E}xtracellular
  {R}ecordings in the {H}ippocampus {I}n {V}ivo.
\newblock {\em J. Neurophysiol.}, 84(1):390--400, 2000.

\bibitem{gray1995tetrodes}
Charles~M Gray, Pedro~E Maldonado, Mathew Wilson, and Bruce McNaughton.
\newblock Tetrodes markedly improve the reliability and yield of multiple
  single-unit isolation from multi-unit recordings in cat striate cortex.
\newblock {\em Journal of neuroscience methods}, 63(1-2):43--54, 1995.

\bibitem{lopez201622}
Carolina~Mora Lopez, Srinjoy Mitra, Jan Putzeys, Bogdan Raducanu, Marco
  Ballini, Alexandru Andrei, Simone Severi, Marleen Welkenhuysen, Chris
  Van~Hoof, Silke Musa, et~al.
\newblock 22.7 a 966-electrode neural probe with 384 configurable channels in
  0.13 $\mu$m soi cmos.
\newblock In {\em Solid-State Circuits Conference (ISSCC), 2016 IEEE
  International}, pages 392--393. IEEE, 2016.

\bibitem{oweiss2007scalable}
Karim~G Oweiss, Andrew Mason, Yasir Suhail, Awais~M Kamboh, and Kyle~E Thomson.
\newblock A scalable wavelet transform vlsi architecture for real-time signal
  processing in high-density intra-cortical implants.
\newblock {\em Circuits and Systems I: Regular Papers, IEEE Transactions on},
  54(6):1266--1278, 2007.

\bibitem{shan2017model}
Kevin~Q Shan, Evgueniy~V Lubenov, and Athanassios~G Siapas.
\newblock Model-based spike sorting with a mixture of drifting t-distributions.
\newblock {\em Journal of neuroscience methods}, 288:82--98, 2017.

\bibitem{pmlr-v70-bojanowski17a}
Piotr Bojanowski and Armand Joulin.
\newblock Unsupervised learning by predicting noise.
\newblock In Doina Precup and Yee~Whye Teh, editors, {\em Proceedings of the
  34th International Conference on Machine Learning}, volume~70 of {\em
  Proceedings of Machine Learning Research}, pages 517--526, International
  Convention Centre, Sydney, Australia, 06--11 Aug 2017. PMLR.

\bibitem{Gibson-SPM-2012}
S.~Gibson, J.~W. Judy, and D.~Markovi$\dot{c}$.
\newblock {S}pike {S}orting: {T}he {F}irst {S}tep in {D}ecoding the {B}rain.
\newblock {\em IEEE Signal Processing Magazine}, 29(1):124--143, 2012.

\bibitem{pachitariu2016fast}
Marius Pachitariu, Nicholas~A Steinmetz, Shabnam~N Kadir, Matteo Carandini, and
  Kenneth~D Harris.
\newblock Fast and accurate spike sorting of high-channel count probes with
  kilosort.
\newblock In {\em Advances in Neural Information Processing Systems}, pages
  4448--4456, 2016.

\bibitem{simonyan2014very}
Karen Simonyan and Andrew Zisserman.
\newblock Very deep convolutional networks for large-scale image recognition.
\newblock {\em arXiv preprint arXiv:1409.1556}, 2014.

\bibitem{sze2017efficient}
Vivienne Sze, Yu-Hsin Chen, Tien-Ju Yang, and Joel~S Emer.
\newblock Efficient processing of deep neural networks: A tutorial and survey.
\newblock {\em Proceedings of the IEEE}, 105(12):2295--2329, 2017.

\bibitem{chen2017Eyeriss}
Yu-Hsin Chen, Tushar Krishna, Joel~S Emer, and Vivienne Sze.
\newblock Eyeriss: An energy-efficient reconfigurable accelerator for deep
  convolutional neural networks.
\newblock {\em IEEE Journal of Solid-State Circuits}, 52(1):127--138, 2017.

\bibitem{xu2014frequency}
Jian Xu, Tong Wu, Wentai Liu, and Zhi Yang.
\newblock A frequency shaping neural recorder with 3 pf input capacitance and
  11 plus 4.5 bits dynamic range.
\newblock {\em IEEE transactions on biomedical circuits and systems},
  8(4):510--527, 2014.

\bibitem{wolf2008thermal}
Patrick~D Wolf.
\newblock Thermal considerations for the design of an implanted cortical
  brain--machine interface (bmi).
\newblock {\em Indwelling Neural Implants: Strategies for Contending with the
  In Vivo Environment}, pages 33--38, 2008.

\end{thebibliography}

\end{document}